\pdfoutput=1
\documentclass[10pt]{article} % For LaTeX2e
\usepackage[accepted]{tmlr}
\usepackage{amsmath,amsfonts,bm}

\def\eqref#1{equation~\ref{#1}}
\def\1{\bm{1}}

\DeclareMathAlphabet{\mathsfit}{\encodingdefault}{\sfdefault}{m}{sl}
\SetMathAlphabet{\mathsfit}{bold}{\encodingdefault}{\sfdefault}{bx}{n}

\usepackage{hyperref}
\usepackage{url}

\usepackage[utf8]{inputenc} % allow utf-8 input
\usepackage[T1]{fontenc}    % use 8-bit T1 fonts
\usepackage{booktabs}       % professional-quality tables
\usepackage{amsfonts}       % blackboard math symbols
\usepackage{nicefrac}       % compact symbols for 1/2, etc.
\usepackage{microtype}      % microtypography
\usepackage{xcolor}         % colors
\usepackage{amsmath}
\usepackage{amssymb}
\usepackage{mathtools}
\usepackage{amsthm}
\usepackage{bbm}
\usepackage{algorithm}
\usepackage{algorithmic}
\usepackage{wrapfig}

\usepackage[capitalize,noabbrev]{cleveref}
\newtheorem{theorem}{Theorem}[section]

\usepackage{colortbl}
\definecolor{mygray}{gray}{.9}
\definecolor{myblue}{rgb}{0.91764706,0.99215686,0.98823529}
\definecolor{none}{rgb}{1,1,1}

\let\AND\relax

\title{Online Continual Learning via Logit Adjusted Softmax}

\author{\name Zhehao Huang \email kinght\_h@sjtu.edu.cn \\
      \addr Institute\ of\ Image\ Processing\ and\ Pattern\ Recognition\\
      Shanghai\ Jiao\ Tong\ University
      \AND
      \name Tao Li \email li.tao@sjtu.edu.cn \\
      \addr Institute\ of\ Image\ Processing\ and\ Pattern\ Recognition\\
      Shanghai\ Jiao\ Tong\ University
      \AND
      \name Chenhe Yuan \email vernunft@sjtu.edu.cn\\
      \addr Department\ of\ Automation \\
      Shanghai\ Jiao\ Tong\ University
      \AND
      \name Yingwen Wu \email yingwen\_wu@sjtu.edu.cn \\
      \addr Institute\ of\ Image\ Processing\ and\ Pattern\ Recognition\\
      Shanghai\ Jiao\ Tong\ University
      \AND
      \name Xiaolin Huang \email xiaolinhuang@sjtu.edu.cn \\
      \addr Institute\ of\ Image\ Processing\ and\ Pattern\ Recognition\\
      Shanghai\ Jiao\ Tong\ University
      \AND
      }

\def\month{05}  % Insert correct month for camera-ready version
\def\year{2024} % Insert correct year for camera-ready version
\def\openreview{\url{https://openreview.net/forum?id=MyQKcQAte6}} % Insert correct link to OpenReview for camera-ready version

\begin{document}

\maketitle

\begin{abstract}
Online continual learning is a challenging problem where models must learn from a non-stationary data stream while avoiding catastrophic forgetting. Inter-class imbalance during training has been identified as a major cause of forgetting, leading to model prediction bias towards recently learned classes. In this paper, we theoretically analyze that inter-class imbalance is entirely attributed to imbalanced class-priors, and the function learned from intra-class intrinsic distributions is the {optimal classifier that minimizes the class-balanced error}. To that end, we present that a simple adjustment of model logits during training can effectively resist prior class bias and pursue the corresponding {optimum}. Our proposed method, Logit Adjusted Softmax, can mitigate the impact of inter-class imbalance not only in class-incremental but also in realistic {scenarios that sum up class and domain incremental learning}, with little additional computational cost. We evaluate our approach on various benchmarks and demonstrate significant performance improvements compared to prior arts. For example, our approach improves the best baseline by 4.6\% on CIFAR10. Codes are available at \url{https://github.com/K1nght/online_CL_logit_adjusted_softmax}.
\end{abstract}
\section{Introduction}
\label{sec:introduction}

Continual learning (CL) has emerged to equip deep learning models with the ability to handle multiple tasks on an unbounded data stream. This paper focuses on the online class-incremental (class-IL)~\citep{DeLange2019ACL} CL problem~\citep{Zhou2023DeepCL}, which holds high relevance to real-world applications~\citep{Wang2022SparCLSC}. In online CL, also known as task-free CL, data is obtained from an unknown non-stationary stream for single-pass training. Class-IL learning, in contrast to task-IL learning,  continuously introduces new classes to the model as the data stream distribution changes, without task-identifiers to assist classification {~\citep{van2022three}}.

\textit{Catastrophic forgetting} {~\citep{McCloskey1989CatastrophicII, Ratcliff1990ConnectionistMO}} is a major obstacle to deploying deep learning models in CL. Recent research attributes catastrophic forgetting to \textit{recency bias}~\citep{chrysakis2023online}, which causes deep neural networks to classify samples into currently learned classes. In fact, this bias in CL is similar to the dominance of head classes in long-tailed distribution learning~\citep{Menon2020LongtailLV}. The vanilla model trained on a long-tailed distribution suffers from inter-class imbalance and tends to infer samples into classes that possess a majority of samples. Previous works~\citep{Ahn2020SSILSS} have also observed that one of the primary causes of catastrophic forgetting is inter-class imbalance throughout training. Growing attention to recency bias and inter-class imbalance has given rise to methods~\citep{Koh2021OnlineCL} to alleviate the negative impact of imbalance, among which recently rehearsal-based methods have been highly successful but still with limitations. Replay buffers will become ineffective for long sequential data streams or tasks with numerous categories. Some methods {~\citep{Guo2022OnlineCL, Prabhu2020GDumbAS}} train only on replayed samples to achieve balanced learning but sacrifice most valuable training data and risk overfitting on the buffer. Methods~\citep{Caccia2021NewIO} that separate gradient updates for old and new classes effectively prevent the impact of imbalance between them but fail to construct clear classification boundaries between old and new classes.

Upon decomposing sample probability in non-stationary data streams through conditional probability ($\text{sample probability} = \text{class-conditional}\times \text{class-prior}$), revealing that recency bias caused by inter-class imbalance is entirely attributable to imbalanced class-priors. The underlying \textit{class-conditional invariant} in online class-IL data streams motivates us to learn a function from intrinsic intra-class distributions instead of traditional sample distributions. We propose Logit Adjusted Softmax (\textbf{LAS}) to resist the impact of class-priors and grasp class-conditionals by simply adjusting the model logits output via input label frequencies in training. Our method is grounded by the  {optimal} classifier that minimizes the class-balanced error in online  {class-IL} setup. Moreover,  {in the challenging online CL scenarios~\citep{Xu2021ClassIncrementalDA} that sum up class- and domain-IL}, we show that preserving knowledge in the class-conditional function can better adapt the learner to changing domains. LAS provides the following three practical benefits in comparison to other previous online CL methods: (1) It can eliminate the prediction bias caused by the imbalance between old and new classes, as well as the inherent inter-class imbalance of the data stream. (2) It is orthogonal to the methods improving replay strategies and plug-in to most of the rehearsal-based methods. (3) It improves performance with nearly no additional computational overhead. 

We evaluate LAS on extensive benchmarks over various datasets and multiple setups. Our LAS lifts the plainest Experience Replay (ER)~\citep{Chaudhry2019ContinualLW} to state-of-the-art performance  {without model expansion or computationally intensive technique}, e.g., improving the accuracy of the best baseline by 4.6\% on CIFAR10~\citep{Krizhevsky2009LearningML} in the online  {class-IL} setup. Furthermore, we notice that inter-class imbalance dominates forgetting in long sequential data streams, which is rarely evaluated and always underestimated in previous work, so we evaluate on the challenging ImageNet~\citep{Deng2009ImageNetAL} and iNaturalist~\citep{Horn2017TheIS}, where our proposed method consistently outperforms previous approaches. In addition to the class-IL setup, LAS also succeeds in the blurry setup and  {and scenarios that sum up class- and domain-IL.}

Key contributions of this paper include: (1) We discover the class-conditional invariant and the  {optimality that minimizes the class-balanced error for} the class-conditional function in online class-IL. (2) We propose eliminating class-priors and learning class-conditionals separately under  {the scenarios that sum up class- and domain-IL.} (3) We introduce to adjust model logit outputs in training with a batch-wise sliding-window estimator for time-varying class-priors to pursue the class-conditional function.
\section{Problem Setup}
\label{sec:problem setup}

Beyond the task-IL setting~\citep{Li2016LearningWF} with clear task-boundaries, we consider a more realistic environment where task-identifiers and task-boundaries are absent at any time, and the total number of labels is unknown. Specifically, let $\mathcal X$ be the instance set and $\mathcal Y$ be the corresponding label set. In online CL, $|\mathcal Y|=\infty$. At time $t\in \mathcal T=\{1,2,\dots\}$, given an unknown non-stationary data stream $\mathcal D_t$ over $\mathcal X \times \mathcal Y$, the learner samples data batch $B_t=\{x_i,y_i\}_{i=1}^{|B_t|}\sim \mathbb P(x,y|\mathcal D_t)$. We refer to $B_t$ as the \textit{incoming batch}. If a pair of instance and label is not stored in the memory, it will be inaccessible in subsequent training unless resampled. 

Commonly, a constrained memory $\mathcal M$ ($|\mathcal M| \leq M$) is utilized to enhance online CL: if the buffer is not empty at time $t$, a \textit{Retrieval} program ensembles several instances and other specific information $I$ to form a \textit{buffer batch} $B_t^{\mathcal M}=Retrieval(B_t,\mathcal M_t)=\{x_i,I_i\}_{i=1}^{|B_t^{\mathcal M}|}\sim \mathbb P(x,I|\mathcal M_t)$. The buffer \textit{Update}s with incoming batches, $\mathcal M_{t+1} \leftarrow Update(B_t, \mathcal M_t)$. Typically, ER~\citep{Chaudhry2019ContinualLW} stores instances and labels $I_i=y_i$, retrievals by random replaying, and updates via reservoir sampling~\citep{Vitter1985RandomSW}. Rehearsal helps to alleviate inter-class imbalance when the number of classes is limited, but can not fundamentally eliminate its impact. The minimum class-prior in memory is bounded by the inverse proportion to the number of observed classes, $\min_{y\in\mathcal Y_t}\mathbb P(y|\mathcal M_t)\leqslant 1/|\mathcal Y_t|\rightarrow 0\ (t\rightarrow\infty)$. When the number of seen classes surges, rehearsal will no longer be able to support balanced inter-class learning.

The learner is a neural network parameterized by $\Theta=\{\theta,w\}$. Function $f_\theta: \mathcal X\rightarrow \mathbb R^D$ extracts feature embeddings with dimension $D$. Following the feature extractor, a single-head linear classifier produces logits, $\Phi(\cdot)=w^\top f_\theta(\cdot):\mathcal X\rightarrow \mathbb R^{|\mathcal Y_t|}$ (for short $\Phi_y(\cdot)=w_y^\top f_\theta(\cdot)$), where $w\in\mathbb R^D\times \mathbb R^{|\mathcal Y_t|}$ represents the weights corresponding to target classes. The dimension of weights in the classifier can grow as more classes have been observed. The learner trains through a surrogate loss averaged on all input instances, $\mathcal L_t:\mathcal Y_t\times \mathbb R^{|\mathcal Y_t|}\rightarrow \mathbb R$ ($\mathcal Y_t$ is the set of all observed labels), typically the softmax cross-entropy loss: 
\begin{equation}\label{eq:softmax ce}
    \mathcal L_{\text{CE}}(y, \Phi(x))
    =-\log{\frac{e^{\Phi_{y}(x)}}{\sum_{y^{\prime} \in\mathcal Y_t} e^{\Phi_{y^{\prime}}(x)}}} 
    =\log [1+\sum_{y^{\prime} \neq y} e^{\Phi_{y^{\prime}}(x)-\Phi_y(x)}].
\end{equation}

\section{Statistical View for Time-varying Distribution Learning}
\label{sec:view}

The standard CL methods learn from the sample probability $\mathbb P(x,y|\rho_t)$ of the target distribution $\rho_t$ (for example $\mathcal D_t$ in practice). The model is encouraged to pursue a posterior probability function $\mathbb P(y|x,\rho_t)$ and to minimize the misclassification error $\mathbb E_{\rho_t}[\mathbb E_{x,y|\rho_t}[y\neq \arg\max_{y'\in \mathcal Y_t}\Phi_{y'}(x)]]$. From Bayesian and conditional probability rule, we notice $\mathbb P(y|x,\rho_t)\propto\mathbb P(x,y|\rho_t)= \mathbb P(x|y, \rho_t)\cdot \mathbb P(y|\rho_t)$, revealing that the sample probability $\mathbb P(x,y|\rho_t)$ of a time-varying distribution is controlled by the class-conditional $\mathbb P(x|y,\rho_t)$ and the class-prior $\mathbb P(y|\rho_t)$. \textit{In unknown non-stationary data streams, inter-class imbalance entirely attributes to time-varying class-priors and is independent of class-conditionals.} Therefore, such a factorization of probability motivates us to learn a class-balanced classifier by exclusively pursuing a \textbf{class-conditional function} $\propto\mathbb P(x|y,\rho_t)$, which is agnostic to arbitrarily imbalanced class-priors. 
The class-conditional function has been widely studied in statistical learning on stable distributions~\citep{Long2017ConditionalAD}. {Similar motivation regarding the decomposition of the sample probability has also been discussed in previous work~\citep{van2021class}. We further extend the analysis of the class-conditional function to non-stationary stream distribution learning and discover the class-balanced optimality of the class-conditional function when learning stream distributions without domain drift, i.e., with fixed class-conditionals, as demonstrated in the following Theorem~\ref{thm:bayes optimal}.}
% The class-conditional function has been widely studied in statistical learning on stable distributions~\citep{Long2017ConditionalAD}, while is first introduced and extended to non-stationary stream distribution learning. In fact, we discover the class-balanced Bayes-optimality of the class-conditional function when learning stream distributions without domain drift, i.e., with fixed class-conditionals, as demonstrated in the following Theorem~\ref{thm:bayes optimal}.

\begin{theorem}\label{thm:bayes optimal}
For the time-varying distribution $\rho_t$, given that its class-conditionals keep the same throughout time, i.e., $\forall t,\mathbb P(x|y,\rho_t)=\mathbb P(x|y,\rho_0)$, the class-conditional function satisfies the   {optimal} classifier $\Phi^*_t$ that minimizes the class-balanced error,
\begin{equation}\label{eq:class-il bayes optimal} 
\Phi_t^*\in \mathop{\arg\min}_{\Phi:\mathcal X\rightarrow \mathbb R^{|\mathcal Y_t|}} \operatorname{CBE}(\Phi,\mathcal Y_t),    \quad     \mathop{\arg\max}_{y \in|\mathcal Y_t|} \Phi_{t,y}^*(x) = \mathop{\arg\max}_{y \in|\mathcal Y_t|} \mathbb{P}(x|y,\rho_t).
\end{equation}
\begin{equation}\label{eq:cbe}
\operatorname{CBE}(\Phi,\mathcal Y_t) = \frac{1}{|\mathcal Y_t|} \sum\limits_{y \in \mathcal Y_t} \mathbb E_{\rho_t}[\mathbb{E}_{x|y,\rho_t}[y \neq \mathop{\arg\max}\limits_{y^{\prime} \in \mathcal Y_t} \Phi_{y^{\prime}}(x)]]. \end{equation}
\end{theorem}

In other words, the   {optimal} class-balanced estimate is the class under which the sample is most likely to appear. $\operatorname{CBE}(\Phi,\mathcal Y_t)$ is the Class-Balanced Error~\citep{Menon2013OnTS} on the current label set $\mathcal Y_t$, extended from the misclassification error for class-balanced evaluation, formally in Equation~\ref{eq:cbe}. {Noting that the class-conditional function corresponds to the Bayes optimal classifier when class-priors are uniform. However, in scenarios where the class-priors are inherently imbalanced, the Bayes optimal classifier fails to minimize the class-balanced error~\citep{Menon2013OnTS}.} Bias towards the most recently occurring classes does not aid in reducing the class-balanced error, but approximation towards real underlying class-conditionals helps balanced classification because the class-balanced error is averaged from the per-class error rate. \textit{Therefore, to address the impact of inter-class imbalance and leverage knowledge from intra-class intrinsic distributions, we propose eliminating class-priors and constructing a class-conditional function in online CL.} The proof of Theorem~\ref{thm:bayes optimal} is in Appendix~\ref{sec:proof}. Following, we discuss two distinct CL scenarios on the critical condition of class-conditionals.

\textbf{Discussion on online class-IL with time-invariant class-conditionals.}
Prior works~\citep{chrysakis2023online} have typically assumed no occurrence of domain drift during the learning process in online class-IL. Although domain drift should be taken into account in realistic scenarios, nearly time-invariant class-conditionals are genuinely feasible in practical situations. For instance, acting as a lifelong species observer in the wild, the agent can find that the target class-conditionals conform to their natural distributions, determined by their semantic information and occurrence frequencies. Without intentional human interference, the concept of natural semantics will remain almost unchanged over a prolonged time, i.e., $\forall t,\mathbb P(x|y,\rho_t)\approx\mathbb P(x|y,\rho_0)$. In the experiments, we mainly adhere to the conventional class-IL configuration of no consideration of domain drift and focus on addressing the issues of inter-class imbalance and forgetting induced by recency bias.

{\textbf{Discussion on online CL scenarios with time-varying class-conditionals.}
In the challenging online CL scenarios where $\mathbb P(y|\rho_t)$ and $\mathbb P(x|y,\rho_t)$ fluctuate as the data stream flows, both inter-class imbalance and intra-class domain drift are crucial considerations. While this setting has been studied in offline incremental setups~\citep{Xie2022GeneralIL}, there has been no research on this topic under online conditions, to the best of our knowledge. We now present our contribution to bridging this gap. In this online CL setup, we eliminate class-priors and focus on the class-conditional function, which should not favor any specific domain but should blend all observed domains uniformly for optimal decision-making,}
\begin{equation}\label{eq:general bayes optimal}
    \mathop{\arg\max}_{y \in|\mathcal Y_t|} \Phi_{t,y}^*(x) = \mathop{\arg\max}_{y \in|\mathcal Y_t|}\frac{1}{t}\sum_{i=1}^t \mathbb{P}(x|y,\rho_i).
\end{equation}
Since previous distributions are unavailable in CL, determining the optimal uniform domain distribution is intractable. Nevertheless, the disparity between the   {optimal classifier that minimizes the class-balanced error} and the learned class-conditional function could be measured by the similarity between their underlying intra-class distributions. We combine with knowledge distillation techniques {~\citep{Li2016LearningWF,Tao2020TopologyPreservingCL,kang2022class,dong2023heterogeneous}} to narrow that disparity in probability space. Results in \S\ref{subsec:gains} show that preserving the knowledge in the class-balanced class-conditional function after eliminating class-priors can better adapt to domain drift than the standard posterior function. Therefore, our proposal is scalable to   {online settings that sum up class- and domain-IL.} Furthermore, it paves the way for developing further efficient solutions for this online scenario by minimizing the intra-class probabilities gap from the optimal class-conditional function, which we intend to explore in future research.
\section{Method}
\label{sec:method}

\subsection{Logit Adjustment Technique}
\label{subsec:logit adjustment}

Now our objective becomes excluding class-priors and establishing an estimator for current class-conditionals, i.e., $\Phi_t:\mathcal X\rightarrow \mathbb R^{|\mathcal Y_t|}$, $\exp(\Phi_{t,y}) \propto \mathbb P(x|y, \rho_t)$. However, it is notoriously difficult to model the class-conditionals explicitly {~\citep{van2021class, zajkac2023prediction}}. To detour this problem, we draw on the Logit Adjustment technique proposed by \citep{Menon2020LongtailLV}. Suppose the optimum scorer obtained by minimizing misclassification error on the target distribution $\rho_t$ at time $t$ is $s_t^*:\mathcal X\rightarrow \mathbb R^{|\mathcal Y_t|}$, $\exp(s_{t,y}^*) \propto \mathbb P(y|x,\rho_t)$. Recalling $\mathbb P(y|x,\rho_t) \propto \mathbb P(x|y, \rho_t)\cdot \mathbb P(y|\rho_t)$, we can derive the relationship between the class-conditional estimator $\Phi_t$ and the optimum scorer $s_t^*$ as follows: 
\begin{equation}\label{eq:optimal softmax minus class prob}
        \mathop{\arg\max}\limits_{y \in|\mathcal Y_t|} e^{{s_{t,y}^*(x)}}
        =\mathop{\arg\max}\limits_{y \in|\mathcal Y_t|} \left(e^{{\Phi_{t,y}(x)}} \cdot \mathbb{P}(y|\rho_t)\right)
        =\mathop{\arg\max}\limits_{y \in|\mathcal Y_t|} \left(\Phi_{t,y}(x)+\ln \mathbb{P}(y|\rho_t)\right).
\end{equation}

Equation~\ref{eq:optimal softmax minus class prob} induces a straightforward method to approximate class-conditionals and to achieve a class-balanced classifier: adjusting the model logits output according to class-priors $\mathbb P(y|\rho_t)$ and directly optimizing the softmax cross-entropy loss.

\subsection{Logit Adjusted Softmax Cross-entropy Loss}
\label{subsec:logit adjusted loss}

We now show how to incorporate Logit Adjustment technique into the softmax cross-entropy loss for the aim of addressing the inter-class imbalance issues in online CL. The modified \textbf{Logit Adjusted Softmax cross-entropy loss} is defined as follows: 
\begin{equation}\label{eq:logit adjusted softmax ce}
    \mathcal L_{\text{LAS}}(y, \Phi(x))
    =-\log \frac{e^{\Phi_y(x)+\tau \cdot \log \pi_{y,t}}}{\sum_{y^{\prime} \in\mathcal Y_t} e^{\Phi_{y^{\prime}}(x)+\tau \cdot \log \pi_{y^{\prime},t}}}
    =\log [1+\sum_{y^{\prime} \neq y}\left(\frac{\pi_{y^{\prime},t}}{\pi_{y,t}}\right)^\tau \cdot e^{\left(\Phi_{y^{\prime}}(x)-\Phi_y(x)\right)}],
\end{equation}
where $\tau$ is the temperature scalar, and $\pi_{y,t}$ is the class prior $\mathbb P(y|\mathcal S_t)$ at time $t$. In practice, $\mathcal S_t$ represents the data point collection from which the model samples input batch each time. Due to the uncertainty of $\mathcal S_t$, it is impossible to pinpoint class priors at each moment. To overcome this barrier, the following \S\ref{subsec:estimator} will provide a simple yet effective method for estimating class-priors in the flowing input stream. Applied to rehearsal-based methods, $\mathcal L_{\text{LAS}}$ will act both on incoming and buffer batches to fully exploit input data. The right-hand side of Equation~\ref{eq:logit adjusted softmax ce} illustrates its distinction to the cross-entropy loss in Equation~\ref{eq:softmax ce}, enforcing a large relative margin between the major class and the minor class, i.e., $(\pi_{\text{major},t}/\pi_{\text{minor},t})^\tau>1$ {~\citep{cao2019learning,tan2020equalization,Menon2020LongtailLV}}.

\subsection{Estimator for Time-varying Class-priors}
\label{subsec:estimator}

In stationary distribution, the Logit Adjustment technique~\citep{Collell2016RevivingTA} can determine class-priors based on a large amount of training data. But when facing an unknown time-varying input data stream, it is required to continuously estimate class-priors $\pi_{y,t}$ for $\mathcal L_{\text{LAS}}$ at each time $t$. Therefore, we propose an intuitive batch-wise estimator with sliding window, where \textit{the occurrence frequency of a label in input batches covered by the sliding time window approximates the corresponding class-prior to that label}. Given the length $l>0$ of the time frame, $\pi_{y,t}$ is calculated as follows:
\begin{equation}\label{eq:estimator}
    \pi_{y,t}=\frac{\sum_{i=t-l+1}^t \sum_{\{x',y'\}\in B_i^{\mathcal S}}\mathbbm{1}(y'=y)}{\sum_{i=t-l+1}^t |B_i^{\mathcal S}|},
\end{equation}
where $\mathbbm{1}(\cdot)$ is the indicator function of label $y$ and $B_t^{\mathcal S}$ is the input batch sampled from the data point collection $\mathcal S_t$. For rehearsal-based methods, the input batch often consists of the incoming and buffer batch, i.e., $B_t^{\mathcal S}=B_t\cup B_t^{\mathcal M}$. The length $l$ of the time window concerns a sensitivity-stability trade-off~\citep{Nagengast2011RisksensitivityAT} with respect to the estimation of class priors, which we further study in the sensitivity analysis of \S\ref{subsec:ablation studies}. 
% In practice , $l$ can be determined by the changing properties of the target data stream distribution. 
% For distributions that vary more gently, we recommend a larger $l$, and vice versa.

\begin{figure}[tb]
    \centering
    \includegraphics[width=0.49\linewidth]{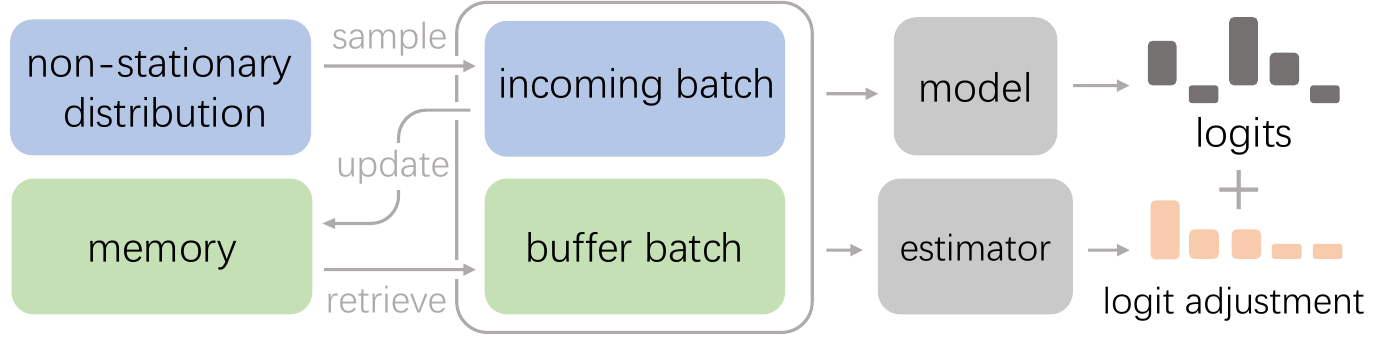}
    \hspace{0.01in}
    \includegraphics[width=0.49\linewidth]{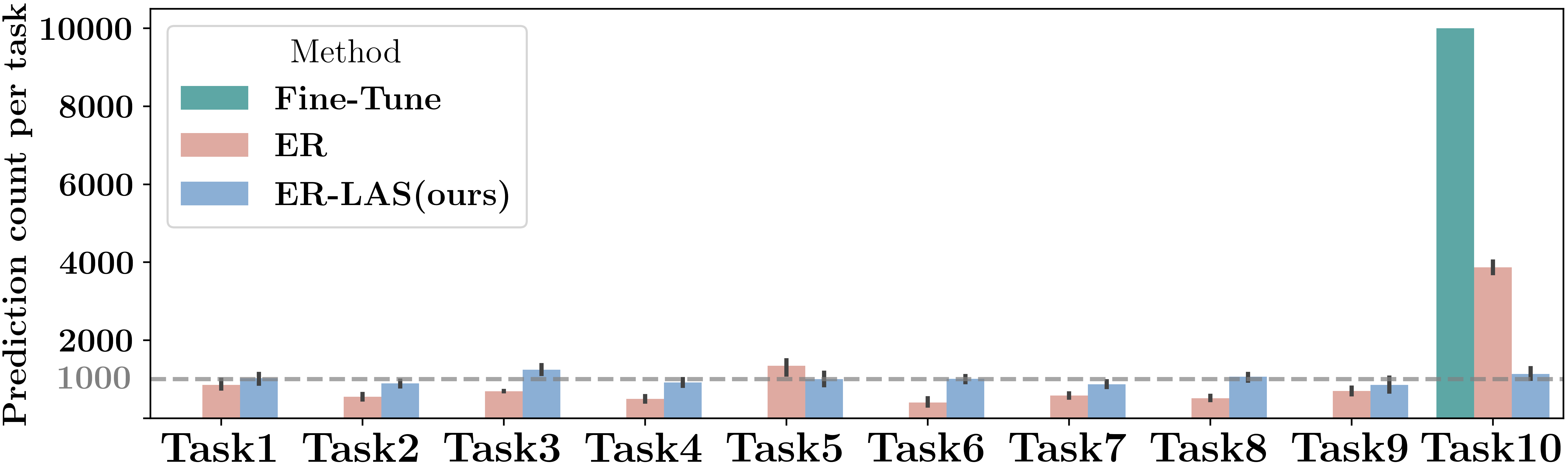}
    \caption{Left is the diagram of Experience Replay (ER) with our proposed Logit Adjusted Softmax and a batch-wise sliding-window estimator (ER-LAS). LAS helps mitigate the inter-class imbalance problem by adding label frequencies to predicted logits. The model in ER-LAS is still trained via the softmax cross-entropy loss. And right is model prediction test samples by Fine-Tune, ER, and ER-LAS on C-CIFAR100 (10 tasks). The gray dashed line indicates the ground truth task-wise distribution ($1k$ for each). We count according to the tasks to which the predicted classes belong.}
    \label{fig:er-las}
\end{figure}

\textbf{Discussion.} Logit Adjusted Softmax cross-entropy loss and the batch-wise estimator with sliding window together constitute our proposed \textbf{LAS} approach. Our method is orthogonal to previous methods of various replay strategies and knowledge distillation techniques. Exact joint label distribution of the non-stationary data stream and the memory retrieval program is unnecessary to our approach, allowing us to effortlessly 
incorporate LAS into existing methods and correct their model prediction bias caused by inter-class imbalance at nearly no cost of additional computational overhead. The experiment in Figure~\ref{fig:er-las}(right) verifies the effect of LAS on correcting the prediction bias, which follows the same setting as in \S\ref{sec:experiment}. Fine-Tune, which trains without any precautions against catastrophic forgetting and inter-class imbalance, categorizes all test samples into the most recently studied task classes. ER~\citep{Chaudhry2019ContinualLW} includes a constrained memory to store previously observed data but still assigns about 38\% (instead of the expected 10\%) test samples to the most recently learned classes. By contrast, our ER-LAS shown in Figure~\ref{fig:er-las}(left) eliminates the recency bias, achieves balanced class-posteriors similar to ground truth distribution, and significantly improves ER performance evaluated in the following \S\ref{sec:experiment}. The algorithm of LAS is in Appendix~\ref{sec:algorithm}.

\textbf{Implementation in online GCL.} We combine LAS with knowledge distillation in online GCL to preserve a class-balanced class-conditional function over averaged domain distributions. We directly calculate the distillation loss between the outputs of old and current models without logit adjustment. Noting that distillation necessitates well-defined task-boundaries to preserve the previous model for distillation. This requirement presents a formidable obstacle in online CL settings, where such boundaries are absent. To investigate the efficacy of our proposed method under the online GCL setting, we allow to acquire task-boundaries in relative experiments. The algorithm of LAS with knowledge distillation for online GCL is {Algorithm}~\ref{alg:kd las} in Appendix~\ref{sec:algorithm}.
\section{Related Work}
\label{sec:related work}

We next provide some intuition on the effectiveness of our proposed approach by comparing LAS to prior work from the perspective of traditional and continual imbalanced distribution learning. We also highlight the computational efficiency in online conditions.

\textbf{Methods for mitigating inter-class imbalance in stable distributions.} \textbf{Logit Adjustment}~\citep{Menon2020LongtailLV} technique appears similar to \textbf{Loss weighting}~\citep{Cui2019ClassBalancedLB} methods, yet the two differ significantly in addressing inter-class imbalance. While Loss weighting methods can balance the representation learning on minority class samples by weighting after the loss between logits and ground truth, it cannot rectify prior class bias and therefore cannot address recency bias. In contrast, Logit Adjustment technique directly balances the class-priors on logits, eradicating the impact of prior class imbalance on model classification and resolving recency bias. In addition to Loss weighting methods, there are also other methods such as \textbf{Weight normalization}~\citep{Kang2019DecouplingRA}, \textbf{Resampling}~\citep{Kubt1997AddressingTC}, and \textbf{Post-hoc correction}~\citep{Collell2016RevivingTA}. Different from these methods and the original Logit Adjustment technique, our adapted LAS possesses firm statistical grounding for non-stationary distributions. We compare with these inter-class imbalance mitigation methods in Appendix~\ref{subsec:more mitigation}.

\textbf{Methods for mitigating inter-class imbalance in non-stationary distributions.} The fundamental \textbf{ER}~\citep{Chaudhry2019ContinualLW} and recently proposed \textbf{ER-ACE}~\citep{Caccia2021NewIO} represent two extreme cases of our approach. ER corresponds to the case where $\tau=0$, and $\mathcal L_{\text{LAS}}$ degenerates into the conventional cross-entropy loss function $\mathcal L_{\text{CE}}$ in Equation~\ref{eq:softmax ce}, losing the ability to alleviate inter-class imbalance. ER-ACE employs asymmetric losses for incoming and buffer batches, considering only the classes present in the current batch for incoming, i.e., $\tau\rightarrow\infty$, and all previously seen classes for replaying, i.e., $\tau=0$, to mitigate representation shift. However, completely separating the gradients of current and past classes blocks the construction of inter-class decision boundaries. Our method lies between ER and ER-ACE, not only pursuing class-conditional function but also encouraging large relative margins between old and new classes in online class-IL, i.e., always $(\pi_{\text{new},t}/\pi_{\text{old},t})^\tau \gg 1$ in Equation~\ref{eq:logit adjusted softmax ce} derived from their imbalance. We also notice highly related \textbf{Logit Rectify} methods~\citep{Zhou2023DeepCL} designed for offline task-IL, which we compare in Appendix~\ref{subsec:more offline}. 

\textbf{Computational efficiency.} Online CL cannot ignore real-time requirements because memory and training time is usually limited in practical scenarios. Compared to traditional Softmax, Logit Adjusted Softmax slightly increases the computational cost of $\mathcal O(|\mathcal Y_t|)$. Our suggested estimator raises the calculation time by $\mathcal O(|B_t|+|B_t^{\mathcal M}|+|\mathcal Y_t|)$ and the memory cost by $\mathcal O(|\mathcal Y_t|)$. In contrast to the time and storage overhead of the model and the memory, such an increase is negligible and lower than in previous works. Our experiments primarily compare methods with computational costs similar to our approach. Noting that CL methods based on contrastive learning {~\citep{Guo2022OnlineCL,guodealing}} may consume substantially more computational resources than our algorithm. We present a performance comparison with these methods in Appendix~\ref{subsec:more contrastive}.
\section{Experiment}
\label{sec:experiment}

In this section, we conduct comprehensive experiments to demonstrate the effectiveness of our proposed LAS. First, we investigate the performance of LAS in the online class-IL scenario with class-disjoint tasks and in the online blurry CL scenario without clear task-boundaries. Then, we evaluate LAS’s gains on rehearsal-based methods in the online class-IL setup and gains on knowledge distillation approaches in the online CL setup {that sums up class- and domain-IL}. Finally, we study the extreme variants of our method, the necessity of the suggested batch-wise estimator with sliding window, and the hyperparameter sensitivity of our LAS.

\paragraph{Benchmark setups.}
We use 5 image classification datasets combined with 3 kinds of CL setups to form 8 benchmarks. Among datasets, \textbf{CIFAR10}~\citep{Krizhevsky2009LearningML} has 10 classes. \textbf{CIFAR100}~\citep{Krizhevsky2009LearningML} has 100 classes, and they can also be categorized into 20 superclasses with 5 domains. \textbf{TinyImageNet}~\citep{Le2015TinyIV} has 200 classes. \textbf{ImageNet} ILSVRC 2012~\citep{Deng2009ImageNetAL} has 1,000 classes, evaluating method performance on the long sequence data stream. \textbf{iNaturalist} 2017~\citep{Horn2017TheIS} has 5,089 classes. The distribution of images per category in iNaturalist follows the observation frequency of the species in the wild, so the data stream possesses inherent inter-class imbalance. As to CL setups, \textbf{online class-IL (C)}~\citep{Aljundi2019OnlineCL} splits a dataset into multiple tasks with uniform disjoint classes, e.g., C-CIFAR10 (5 tasks) is split into 5 disjoint tasks with 2 classes each, except for C-iNaturalist (26 tasks) that is organized into 26 disjoint tasks according to the initial letter of each class. \textbf{Online blurry CL (B)}~\citep{Koh2021OnlineCL} has both class-IL distributions and blurry task boundaries. It divides the classes into $N_{\text{blurry}}\%$ disjoint part and $(100-N_{\text{blurry}}\%)$ blurry part. The disjoint part classes only appear in fixed tasks, while the blurry part classes occur throughout the data stream but with inherent inter-class imbalance represented by blurry level $M_{\text{blurry}}$. We split CIFAR100 and TinyImageNet into 10 blurry tasks according to \citep{Koh2021OnlineCL} with disjoint ratio $N_{\text{blurry}}=50$ and blurry level $M_{\text{blurry}}=10$. {\textbf{Online Sum-Class-Domain CL (S)}~\citep{Xie2022GeneralIL} covers class- and domain-IL setup, where incoming data contains images from new classes and new domains. We only apply this online CL setup on CIFAR100.} The learner needs to predict superclass labels. Each superclass has 5 subclasses representing 5 different domains within the same class. See Appendix~\ref{sec:benchmark} for more details about benchmark setups.

\textbf{Training Protocol.}
For all experiments, unless otherwise specified, following \citep{Buzzega2020DarkEF}. We use the full ResNet18 as the feature extractor. For small-scale datasets, we start training from scratch. We pre-train models on 100 randomly selected classes from C-ImageNet and then perform online learning on the remaining 900 classes\citep{Gallardo2021SelfSupervisedTE}. As for C-iNaturalist, we pre-train models on the entire ImageNet dataset. A single-head classifier is applied to classify all seen labels. We use SGD optimizer without momentum and weight decay. The learning rate is set to 0.03 and kept constant. Incoming and buffer batch sizes are both 32. On C-ImageNet and C-iNaturalist, we set both batch sizes to 128. We apply standard data augmentation, including \textit{random-resized-crop}, \textit{horizontal-flip}, and \textit{normalization}. Some literature\citep{Koh2021OnlineCL} assumes that data arrive one-by-one in online CL, in which case we can accumulate samples as a batch to help model optimization convergence. We discuss the performance under varying batch sizes and per-sample updating in Appendix~\ref{subsec:more batch}. For online CL, only one epoch is used to run all methods for each task, and gradient descent is performed only once per incoming batch. By default, we set $\tau=1.0$ and $l=1$ for LAS. We report means and standard deviations of all results across 10 independent runs.

\textbf{Evaluation Protocol.} 
A commonly used metric is the final average accuracy $A_T$. Another common metric is the final average forgetting~\citep{Chaudhry2020ContinualLI} $F_T$. For blurry setup, we follow \citep{Koh2021OnlineCL} to add the Area Under the Curve of Accuracy $A_{\text{AUC}}$ to evaluate the model performance throughout training. The detailed computation of each metric is given in Appendix~\ref{sec:metrics}. 

\textbf{Baselines.}
% We consider 7 rehearsal-based methods for online CL to compare: ER is the most fundamental replay method, which updates via a reservoir and randomly replays samples and labels. DER++ replays samples with cross-entropy loss over ground truth labels and distillation loss over logits. MRO also uses a reservoir to update memory but only trains from memory. SS-IL separates the cross-entropy of the observed classes from the absent classes. CLIB is intended for online class-IL blurry data streams, calculates sample-wise importance to update class-balanced memory, and only trains on replayed samples. ER-ACE employs asymmetric loss, which only considers the occurred classes for the incoming batch and all seen classes for replayed samples. ER-OBC additionally freezes the feature extractor and only updates the classifier layer by a new buffer batch after each step. We enhance 3 methods of replay strategy: MIR improves the retrieval program by prioritizing the memory samples most interfered with by the model updating. ASER$_{\mu}$ calculates Shapley values of samples to update and retrieve. OCS selects a representative and diverse coreset with a high affinity to train and replay. Also, 3 approaches of knowledge distillation are augmented by LAS: LwF computes distillation loss on logits output of previous classes from the old network. LUCIR proposes distillation loss on normalized features from the old feature extractor. GeoDL proposes to also distill in the feature space but measure by the geodesic path.
We consider 7 rehearsal-based methods for online CL to compare: \textbf{ER}~\citep{Chaudhry2019ContinualLW} uses reservoir update and random replay. 
\textbf{DER++}~\citep{Buzzega2020DarkEF} replays samples with previous logits for distillation loss. 
\textbf{MRO}~\citep{chrysakis2023online} only trains from memory. 
\textbf{SS-IL}~\citep{Ahn2020SSILSS} separates the loss for present and absent classes. 
\textbf{CLIB}~\citep{Koh2021OnlineCL} updates by sample-wise importance and only trains on replayed samples. 
\textbf{ER-ACE}~\citep{Caccia2021NewIO} employs the asymmetric loss to reduce representation shift. 
\textbf{ER-OBC}~\citep{chrysakis2023online} additionally updates the classifier by balanced buffer batches. 
\textbf{ER-CBA}~\citep{wang2023cba} introduces a continual bias adapter inserted after the classifier and conducts dual optimization on input and buffer batches.
In addition, we enhance 3 methods of replay strategy: 
\textbf{MIR}~\citep{Aljundi2019OnlineCL} retrieves the memory samples most interfered with by the model updating. 
\textbf{ASER}$_{\mu}$~\citep{Shim2020OnlineCC} calculates Shapley values of samples to update and retrieve. 
\textbf{OCS}~\citep{yoon2022online} selects coreset with high affinity to replay. Also, knowledge distillation losses in 3 approaches are augmented by LAS: 
\textbf{LwF}~\citep{Li2016LearningWF} distills on logits of previous classes. 
\textbf{LUCIR}~\citep{Hou2019LearningAU} distills on normalized features. 
\textbf{GeoDL}~\citep{Simon2021OnLT} also distills in the feature space but measures by the geodesic path. 

{
\begin{table*}[t]
    \caption{Final average accuracy $A_T$ (higher is better) on C-CIFAR10 (5 tasks), C-CIFAR100 (10 tasks), and C-TinyImageNet (10 tasks). $M$ is memory size.}
    \label{tab:class cl}
    \small
    \centering
    \resizebox{\columnwidth}{!}{
    \begin{tabular}{@{}l | ccc | ccc | ccc@{}} 
    \toprule
    \textbf{Dataset}         &\multicolumn{3}{c|}{C-CIFAR10}&\multicolumn{3}{c|}{C-CIFAR100}& \multicolumn{3}{c}{C-TinyImageNet}\\
    \midrule
    \textbf{Method}          & $M=0.5k$ & $M=1k$ & $M=2k$   & $M=0.5k$ & $M=1k$ & $M=2k$   & $M=0.5k$ & $M=1k$ & $M=2k$ \\
    % \midrule
    \midrule
    ER              &$40.9$\tiny{$\pm1.2$}&$45.4$\tiny{$\pm1.8$}&$50.3$\tiny{$\pm1.1$}  &$12.9$\tiny{$\pm0.3$}&$16.5$\tiny{$\pm0.4$}&$19.8$\tiny{$\pm0.6$}  &$8.8$\tiny{$\pm0.2$}&$11.0$\tiny{$\pm0.2$}&$14.3$\tiny{$\pm0.3$}\\
    DER++           &$49.4$\tiny{$\pm1.0$}&$49.7$\tiny{$\pm3.0$}&$48.9$\tiny{$\pm0.9$}  &$8.9$\tiny{$\pm0.4$}&$13.1$\tiny{$\pm0.4$}&$12.3$\tiny{$\pm0.4$}   &$5.9$\tiny{$\pm0.2$}&$8.0$\tiny{$\pm0.3$}&$9.5$\tiny{$\pm0.3$}\\
    MRO             &$43.4$\tiny{$\pm1.0$}&$49.3$\tiny{$\pm1.1$}&$55.9$\tiny{$\pm0.6$}  &$11.5$\tiny{$\pm0.1$}&$18.3$\tiny{$\pm0.2$}&$23.1$\tiny{$\pm0.1$}  &$5.9$\tiny{$\pm0.1$}&$9.2$\tiny{$\pm0.1$}&$13.4$\tiny{$\pm0.2$}\\
    SS-IL           &$47.7$\tiny{$\pm0.7$}&$52.6$\tiny{$\pm0.5$}&$51.7$\tiny{$\pm0.4$}  &$19.2$\tiny{$\pm0.2$}&$21.5$\tiny{$\pm0.2$}&$24.2$\tiny{$\pm0.2$}  &$13.1$\tiny{$\pm0.2$}&$14.9$\tiny{$\pm0.1$}&$17.1$\tiny{$\pm0.9$}\\
    CLIB            &$48.4$\tiny{$\pm0.9$}&$54.8$\tiny{$\pm1.0$}&$55.9$\tiny{$\pm1.0$}  &$15.9$\tiny{$\pm0.2$}&$20.7$\tiny{$\pm0.2$}&$25.3$\tiny{$\pm0.3$}  &$8.3$\tiny{$\pm0.1$}&$12.1$\tiny{$\pm0.2$}&$15.9$\tiny{$\pm0.2$}\\
    ER-ACE          &$44.4$\tiny{$\pm1.0$}&$48.1$\tiny{$\pm1.1$}&$51.2$\tiny{$\pm1.2$}  &$18.6$\tiny{$\pm0.4$}&$22.5$\tiny{$\pm0.5$}&$25.0$\tiny{$\pm0.9$}  &$11.4$\tiny{$\pm0.2$}&$14.8$\tiny{$\pm0.2$}&$16.4$\tiny{$\pm0.4$}\\
    ER-OBC          &$45.1$\tiny{$\pm0.6$}&$46.4$\tiny{$\pm0.6$}&$46.0$\tiny{$\pm0.4$}  &$15.6$\tiny{$\pm0.2$}&$17.9$\tiny{$\pm0.2$}&$22.1$\tiny{$\pm0.3$}  &$9.1$\tiny{$\pm0.1$}&$13.2$\tiny{$\pm0.1$}&$16.4$\tiny{$\pm0.1$}\\
    ER-CBA          &{$45.0$\tiny{$\pm1.6$}}&{$54.2$\tiny{$\pm1.1$}}&{$56.3$\tiny{$\pm0.9$}}  &{$\mathbf{20.1}$\tiny{$\mathbf{\pm0.6}$}}&{$23.0$\tiny{$\pm0.3$}}&{$26.0$\tiny{$\pm0.6$}}  &{$12.3$\tiny{$\pm0.7$}}&{$13.6$\tiny{$\pm0.5$}}&$17.0$\tiny{$\pm0.5$}\\
    \rowcolor{myblue}
    ER-LAS          &$\mathbf{51.7}$\tiny{$\mathbf{\pm0.9}$}&$\mathbf{55.3}$\tiny{$\mathbf{\pm1.6}$}&$\mathbf{60.5}$\tiny{$\mathbf{\pm0.8}$}  &$\mathbf{20.1}$\tiny{$\mathbf{\pm0.2}$}&$\mathbf{25.7}$\tiny{$\mathbf{\pm0.3}$}&$\mathbf{27.0}$\tiny{$\mathbf{\pm0.3}$}  &$\mathbf{13.7}$\tiny{$\mathbf{\pm0.2}$}&$\mathbf{15.5}$\tiny{$\mathbf{\pm0.2}$}&$\mathbf{18.7}$\tiny{$\mathbf{\pm0.2}$}\\
    \bottomrule
    \end{tabular}}
\end{table*}
}

{
\begin{table}[t]
\begin{tabular}{cc}
\hspace{-1.0em}
  \begin{minipage}{0.49\textwidth}
  \centering
      \makeatletter\def\@captype{table}\makeatother\caption{Final average accuracy $A_T$ (higher is better) and final average forgetting $F_T$ (lower is better) on C-ImageNet (90 tasks) and C-iNaturalist (26 tasks). We show the results of top-3 methods. Memory sizes are $M=20k$.}
      \label{tab:imagenet-inat}
      \resizebox{0.97\columnwidth}{!}{
      \begin{tabular}{@{}l|c|c@{}}
      \toprule
      \textbf{Dataset} & C-ImageNet                      & C-iNaturalist \\
      \midrule
      \textbf{Method}  & $A_T\uparrow$ / $F_T\downarrow$         & $A_T\uparrow$ / $F_T\downarrow$ \\
      \midrule
      ER & \small$31.8$\tiny{$\pm0.1$} \small{$/$} \small$38.6$\tiny{$\pm0.2$}                        & \small$4.7$\tiny{$\pm0.0$} \small{$/$} \small$18.0$\tiny{$\pm0.0$}  \\
      ER-ACE & \small$33.4$\tiny{$\pm0.2$} \small{$/$} \small$11.3$\tiny{$\pm0.1$}                    & \small$5.7$\tiny{$\pm0.0$} \small{$/$} \small$1.1$\tiny{$\pm0.0$}  \\ 
      MRO & \small$35.8$\tiny{$\pm0.1$} \small{$/$} \small$10.2$\tiny{$\pm0.2$}                     & \small$5.0$\tiny{$\pm0.0$} \small{$/$} \small$\bf{0.4}$\tiny{$\bf{\pm0.0}$}  \\
      \rowcolor{myblue}
      ER-LAS & \small$\bf{39.3}$\tiny{$\bf{\pm0.1}$} \small{$/$} \small$\bf{9.0}$\tiny{$\bf{\pm0.1}$}                     & \small$\bf{8.1}$\tiny{$\bf{\pm0.0}$} \small{$/$} \small$2.8$\tiny{$\pm0.0$}  \\
      \bottomrule
\end{tabular}}
  \end{minipage}
  & 
  \begin{minipage}{0.49\textwidth}
    \centering
        \makeatletter\def\@captype{table}\makeatother\caption{AUC of Accuracy $A_{\text{AUC}}$ and final average accuracy $A_T$ (both higher is better) on B-CIFAR100 (10 tasks) and B-TinyImageNet (10 tasks). We show the results of top-3 methods. Memory sizes are $M=2k$.}
        \label{tab:blurry cl}
        \resizebox{\columnwidth}{!}{
        \begin{tabular}{@{}l | c | c@{}} 
        \toprule
        \textbf{Dataset}             & B-CIFAR100                   & B-TinyImageNet\\
        \midrule
        \textbf{Method}              &  $A_{T}\uparrow$ / $A_{\text{AUC}}\uparrow$   & $A_{T}\uparrow$ / $A_{\text{AUC}}\uparrow$ \\
        % \midrule
        \midrule
        ER   &\small$19.6$\tiny{$\pm1.6$}\small{$/$}\small$16.1$\tiny{$\pm0.1$} &\small$16.2$\tiny{$\pm0.2$}\small{$/$}\small$12.4$\tiny{$\pm0.0$}\\
        ER-ACE       &\small$18.3$\tiny{$\pm1.0$}\small{$/$}\small$15.2$\tiny{$\pm0.0$} &\small$16.4$\tiny{$\pm0.3$}\small{$/$}\small$12.2$\tiny{$\pm0.1$}\\
        CLIB         &\small$21.9$\tiny{$\pm0.3$}\small{$/$}\small$18.0$\tiny{$\pm0.1$} &\small$15.9$\tiny{$\pm0.2$}\small{$/$}\small$12.6$\tiny{$\pm0.1$}\\
        \rowcolor{myblue}
        ER-LAS              &\small$\mathbf{24.9}$\tiny{$\mathbf{\pm0.5}$}\small{$/$}\small$\mathbf{20.3}$\tiny{$\mathbf{\pm0.0}$} &\small$\mathbf{19.4}$\tiny{$\mathbf{\pm0.4}$}\small{$/$}\small$\mathbf{15.1}$\tiny{$\mathbf{\pm0.0}$} \\
        \bottomrule
        \end{tabular}}
    \end{minipage}
\end{tabular}
\end{table}
}

\subsection{Results on Online Class-IL Scenarios}
\label{subsec:online class-il cl results}

\textbf{Accuracy results.}
Table~\ref{tab:class cl} and Table~\ref{tab:imagenet-inat} show the final average accuracy for C-CIFAR10, C-CIFAR100, C-TinyImageNet, C-ImageNet, and C-iNaturalist with various memory sizes. ER-LAS consistently outperforms all compared baselines, achieving 60.5\% (+4.6\%), 27.0\% (+1.7\%), and 18.7\% (+1.6\%) on C-CIFAR10, C-CIFAR100, C-TinyImageNet respectively compared to the best baselines. Compared to only considering replayed samples in MRO and CLIB or separating the gradients between old and new classes in SS-IL and ER-ACE, LAS optimizes for a class-balanced function for incoming and buffer batches and enforces large relative margins between imbalanced classes, resulting in better performance. Considering that the challenging C-ImageNet and C-iNaturalist benchmarks possess substantially longer sequences of data stream than the above three benchmarks, where the recency bias problem caused by inter-class imbalance becomes severely critical, we also apply LAS to boost the performance of ER. We present the results of the top-3 baselines (MRO, ER-ACE, ER) on C-ImageNet and C-iNaturalist. ER-LAS can obtain 39.3\% (+3.5\%) on C-ImageNet and 8.1\% (+2.4\%) on C-iNaturalist compared to the best baselines. {To ensure a fair comparison on C-iNaturalist, we present both the final average accuracy directly evaluated on the test data like other methods do, and the results of the final average class-balanced accuracy aligning with our \cref{thm:bayes optimal} in Appendix F.7.} Our extensive evaluations demonstrate the superior performance of our LAS by effectively alleviating inter-class imbalance in the online class-IL setup with nearly no additional computation cost (Table~\ref{tab:training time}). ER-LAS is only slightly slower than ER, contributing to its real-world online applications.

\textbf{Forgetting rate.} We compare the final average forgetting of ER-LAS with top-3 performed baselines (MRO, ER-ACE, ER) on C-ImageNet and C-iNaturalist. As shown in Table~\ref{tab:imagenet-inat}, ER-LAS achieves the least forgetting rate on C-ImageNet and only forgets more than MRO and ER-ACE on C-iNaturalist. However, the lowest forgetting rate (e.g., 0.4\% of MRO) does not necessarily guarantee the highest accuracy (8.1\% of ER-LAS) because of the stability-plasticity dilemma~\citep{Kim2023OnTS}. In the following sensitivity analysis of \S~\ref{subsec:ablation studies}, we show that although a lower forgetting rate can be obtained by deliberately tuning hyperparameters in our LAS, a better stability-plasticity trade-off can be achieved by the optimal hyperparameters. It is worth noting that in long sequence benchmarks, compared to ER without considering inter-class imbalance, methods trying to address recency bias not only remarkably reduce forgetting rates but also bring about improvements in accuracy, underscoring the importance of inter-class imbalance as a top priority in lifelong class-IL. We provide prediction results on C-ImageNet to further support our efficacy of eliminating recency bias in Appendix~\ref{subsec:more predict}. We also evaluate the final average forgetting on C-CIFAR10, C-CIFAR100, and C-TinyImageNet in Appendix~\ref{subsec:more forget}.

{
\begin{table}[t]
\begin{tabular}{cc}
\hspace{-1.0em}
  \begin{minipage}{0.49\textwidth}
    \centering
        \makeatletter\def\@captype{table}\makeatother\caption{Final average accuracy $A_T$ (higher is better) by replay strategy methods w/o and w/ LAS on C-CIFAR100 (10 tasks). Gains are shown in parentheses. $M$ is memory size.}
        \label{tab:rehearsal gain}
        \resizebox{\columnwidth}{!}{
        \begin{tabular}{@{}l|cc@{}}
        \toprule
        \textbf{Dataset}  & \multicolumn{2}{c}{C-CIFAR100} \\
        \midrule
        \textbf{Method}   &$M=0.1k$ &$M=0.5k$ \\
        \midrule
        ER                &$6.5$\tiny{$\pm0.2$} &$12.9$\tiny{$\pm0.3$}  \\
        \rowcolor{myblue}
        ER-LAS            &$10.7$\tiny{$\pm0.2$} \normalsize($4.2\uparrow$) &$20.1$\tiny{$\pm0.2$} \normalsize($7.2\uparrow$)  \\
        \midrule
        \rowcolor{none}
        MIR               &$6.6$\tiny{$\pm0.3$} &$12.0$\tiny{$\pm0.3$} \\
        \rowcolor{myblue}
        MIR-LAS           &$11.8$\tiny{$\pm0.1$} \normalsize($5.2\uparrow$) &$21.1$\tiny{$\pm0.2$} \normalsize($9.1\uparrow$)  \\
        \midrule
        \rowcolor{none}
        ASER$_\mu$        &$7.8$\tiny{$\pm0.2$} &$13.8$\tiny{$\pm0.3$} \\
        \rowcolor{myblue}
        ASER$_\mu$-LAS    &$9.5$\tiny{$\pm0.4$} \normalsize($1.7\uparrow$) &$18.0$\tiny{$\pm0.3$} \normalsize($4.2\uparrow$)  \\
        \midrule
        \rowcolor{none}
        OCS               &$9.4$\tiny{$\pm0.1$} &$16.2$\tiny{$\pm0.2$}  \\
        \rowcolor{myblue}
        OCS-LAS           &$12.7$\tiny{$\pm0.2$} \normalsize($3.3\uparrow$) &$21.0$\tiny{$\pm0.3$} \normalsize($4.8\uparrow$)  \\
        \bottomrule
        \end{tabular}}
    \end{minipage}
  & 
  \begin{minipage}{0.49\textwidth}
    \centering
        \makeatletter\def\@captype{table}\makeatother\caption{Final average accuracy $A_T$ (higher is better) by knowledge distillation approaches w/o and w/ LAS on {S-CIFAR100} (20 tasks). Gains are shown in parentheses. $M$ is memory size.}
        \label{tab:general gain}
        \resizebox{\columnwidth}{!}{
        \begin{tabular}{@{}l|cc@{}}
        \toprule
        \textbf{Dataset} & \multicolumn{2}{c}{{S-CIFAR100}} \\
        \midrule
        \textbf{Method}  & $M=0.1k$ & $M=0.5k$ \\
        \midrule
        ER                                        &$20.4$\tiny{$\pm0.2$} &$27.3$\tiny{$\pm0.4$}  \\
        \rowcolor{myblue}
        ER-LAS                                       &$24.1$\tiny{$\pm0.2$} \normalsize($3.7\uparrow$) &$31.5$\tiny{$\pm0.5$} \normalsize($4.2\uparrow$)\\
        \midrule
        \rowcolor{none}
        LwF              &$23.9$\tiny{$\pm0.3$} &$30.1$\tiny{$\pm0.3$} \\
        \rowcolor{myblue}
        LwF-LAS             &$26.0$\tiny{$\pm0.2$} \normalsize($2.1\uparrow$) &$32.4$\tiny{$\pm0.1$} \normalsize($2.3\uparrow$) \\
        \midrule
        \rowcolor{none}
        LUCIR              &$20.1$\tiny{$\pm0.1$} &$29.4$\tiny{$\pm0.3$} \\
        \rowcolor{myblue}
        LUCIR-LAS             &$25.0$\tiny{$\pm0.2$} \normalsize($4.9\uparrow$) &$32.6$\tiny{$\pm0.3$} \normalsize($3.2\uparrow$)  \\
        \midrule
        \rowcolor{none}
        GeoDL            &$20.6$\tiny{$\pm0.2$} &$30.1$\tiny{$\pm0.2$}  \\
        \rowcolor{myblue}
        GeoDL-LAS           &$25.2$\tiny{$\pm0.2$} \normalsize($4.6\uparrow$) &$32.8$\tiny{$\pm0.2$} \normalsize($2.7\uparrow$)  \\
        \bottomrule
        \end{tabular}}
    \end{minipage}
\end{tabular}
\end{table}
}

\subsection{Results on Online Blurry CL Scenarios}
\label{subsec:blurry results}

We compare ER-LAS with the best 3 baselines on B-CIFAR100 and B-TinyImageNet. Table~\ref{tab:blurry cl} shows that ER-LAS can outperform all baselines on $A_T$ and $A_{\text{AUC}}$. For example, ER-LAS improves the best baseline by 3.0\% $A_T$ and 2.5\% $A_{\text{AUC}}$ on B-TinyImageNet. In fact, our method is particularly suitable for the online blurry CL setup because LAS alleviates the detrimental effects of inter-class imbalance both inherently in the data stream and between new and old classes. The results of ER-LAS further confirm that such an advantage can help obtain high accuracy throughout learning under the realistic online blurry CL setup with challenging inter-class imbalance problems.

\subsection{Gains on Enhanced Methods}
\label{subsec:gains}

\textbf{Rehearsal-based methods on online class-IL scenarios.} We verify the performance boost of LAS by plugging it into ER, MIR, ASER$_\mu$, and OCS. These three baselines train via softmax cross-entropy loss with different replay strategies, which harmonize with our approach. Table~\ref{tab:rehearsal gain} shows that LAS can significantly improve ER and its variants (+1.7\%$\sim$+9.1\%) in the online class-IL setup. Although these methods with various memory management strategies benefit from our LAS, the gains depend on the estimation of class-priors from retrieval, as a relatively smaller boost is observed on ASER$_\mu$ which has a sophisticated strategy to manage memory.

\textbf{Knowledge distillation methods on online {Sum-Class-Domain} CL scenarios.} 
To further investigate LAS’s effectiveness in alleviating inter-class imbalance, we combined it with knowledge distillation approaches in the difficult and realistic online CL setup {that sums up class- and domain-IL}. Table~\ref{tab:general gain} summarizes the results. $\mathcal L_{\text{CE}}$ represents the basic CE loss used in ER. Knowledge distillation losses obtain higher accuracy by adapting intra-class domain drift. Augmented by our LAS, consistent gains (+2.1\%$\sim$+4.9\%) are observed by eliminating class-imbalanced prior bias. The results demonstrate the validity of our proposal to separately handle class-conditionals and class-priors in non-stationary stream learning. It also showcases the performance improvement of eliminating imbalanced class-priors by our method in {this} CL setup. Noting that we allowed the knowledge distillation methods to preserve old models at boundaries, which is intractable in real-world online CL. In future studies, we will explore the efficient and task-free method for handling intra-class domain drift to further refine the solution to online  {Sum-Class-Domain} CL. 

{
\begin{table}[t]
\begin{tabular}{cc}
  \hspace{-1.0em}
  \begin{minipage}{0.49\textwidth} 
    \begin{minipage}[h]{\textwidth} 
      \centering
      \makeatletter\def\@captype{table}\makeatother\caption{Ablation study about two extreme situations of $\tau$ and about randomly assigned ($Random$) or macro statistical ($Macro$) class-priors on C-CIFAR100 (10 tasks). $M=2k$.}
      \label{tab:ablation study}
      \resizebox{\columnwidth}{!}{
      \begin{tabular}{@{}l|c|c|c|c|>{\columncolor{myblue}}c@{}}
      \toprule
      \textbf{Method} & $\tau=0$ & $\tau=\infty$ & $Random$ & $Macro$ & LAS \\
      \midrule
      $A_T\uparrow$   &$19.4$\tiny{$\pm0.4$}&$22.7$\tiny{$\pm0.2$}&$20.6$\tiny{$\pm0.2$}&$22.1$\tiny{$\pm0.6$}&$\bf{27.0}$\tiny{$\bf{\pm0.3}$}\\ 
      \midrule
      $F_T\downarrow$ &$29.1$\tiny{$\pm0.4$}&$\bf{2.7}$\tiny{$\bf{\pm0.4}$}&$23.5$\tiny{$\pm0.2$}&$14.2$\tiny{$\pm0.8$}&$10.7$\tiny{$\pm0.4$}\\
      \bottomrule
      \end{tabular}}
    \end{minipage}

    \begin{minipage}[h]{\textwidth} 
      \centering
      \makeatletter\def\@captype{table}\makeatother\caption{Training time compared with top-3 fast methods on C-CIFAR100 (10 tasks) by one Nvidia Geforce GTX 2080 Ti. $M=2k$.}
      \label{tab:training time}
      \resizebox{\columnwidth}{!}{
      \begin{tabular}{@{}l|c|c|c|>{\columncolor{myblue}}c@{}}
      \toprule
      \textbf{Method} & ER & ER-ACE & MRO & ER-LAS \\
      \midrule        
      Training Time (s)   &$77.4$ ($\times 0.94$) &$84.7$ ($\times 1.02$) &$99.0$ ($\times 1.20$) &$82.6$ ($\times 1.00$)\\
      \bottomrule
      \end{tabular}}
    \end{minipage}
  \end{minipage}
  & 
  \begin{minipage}{0.49\textwidth} 
    \centering 
    \includegraphics[width=0.95\textwidth]{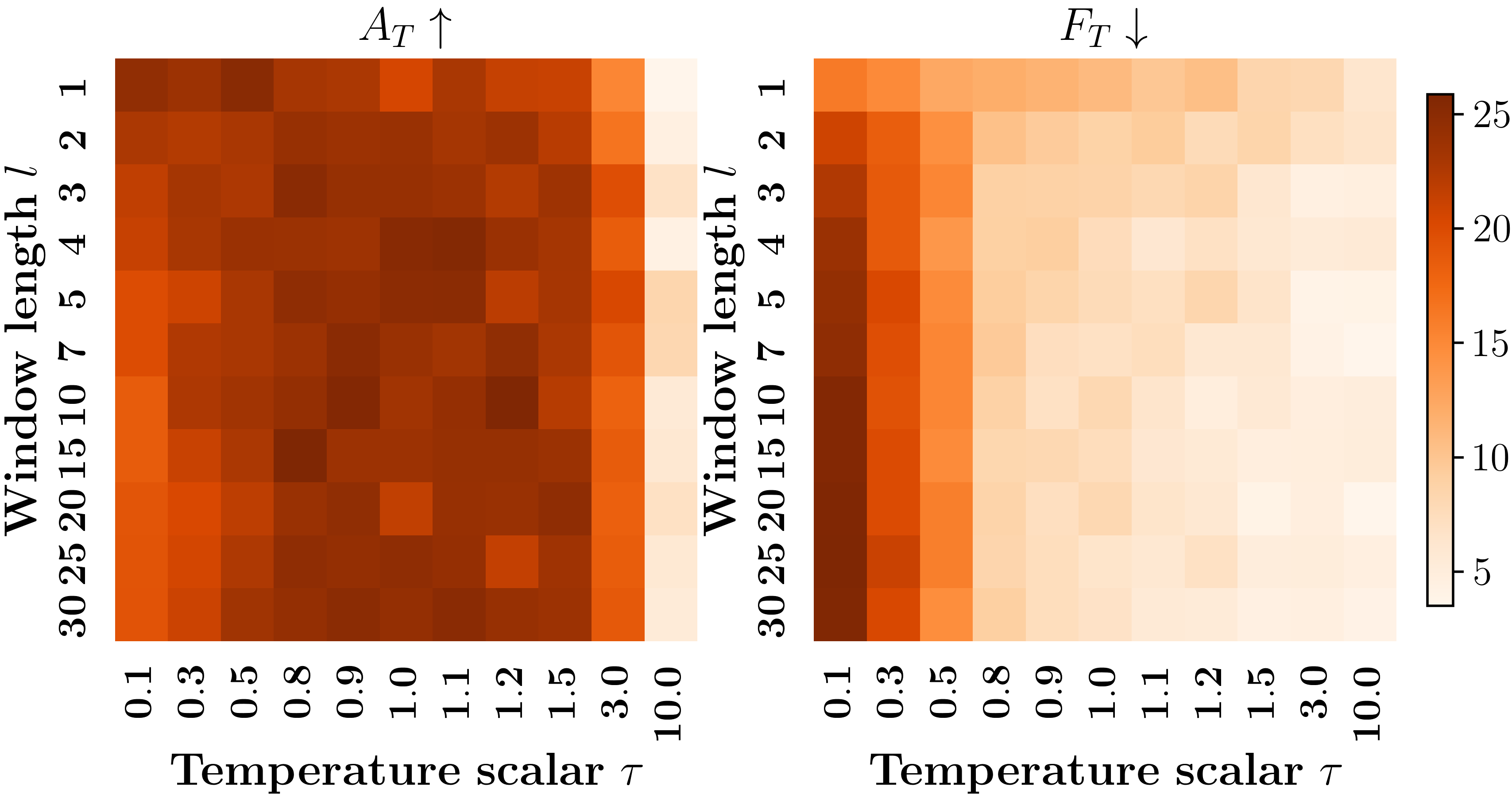} 
    \makeatletter\def\@captype{figure}\makeatother\caption{Final average accuracy (darker is better, left) and final average forgetting (lighter is better, right) of various hyperparameter combinations on C-CIFAR100 (10 tasks). $M=2k$.}
    \label{fig:sensitive}
  \end{minipage}
\end{tabular}
\end{table}
}

\subsection{Ablation Studies}
\label{subsec:ablation studies}

\textbf{Extreme variants of LAS.} 
We investigate the performance of two variants of our method by pushing $\tau$ towards two extremes. When $\tau=0$, LAS degenerates into the traditional softmax cross-entropy loss in ER. In $\tau=\infty$, we set $({\pi_{y^{\prime},t}}/{\pi_{y,t}})^\tau=0$ in Equation~\ref{eq:logit adjusted softmax ce} when ${\pi_{y^{\prime},t}}/{\pi_{y,t}}<1$, otherwise we keep this coefficient and set $\tau=1$ to ensure runnable. It achieves a similar effect as separating the gradient of new and old categories in ER-ACE, reducing representation shift. As shown in Table~\ref{tab:ablation study}, the performance of $\tau=0$ is similar to ER as expected. $\tau=\infty$ benefits from a remarkably low forgetting rate. However, our proposed LAS with $\tau=1$ achieves the highest accuracy, indicating that enforcing a relative margin between classes based on the imbalanced class-priors can obtain a better stability-plasticity trade-off.

\textbf{Necessity of batch-wise estimator with sliding window.}
We empirically validate the necessity of our designed estimator.  We randomly assign each prior of seen classes by a uniform distribution $U[0,1]$ and normalize them to $1$, as $Random$. We also explicitly calculate the joint label distribution of the current data stream and the memory replay, as $Macro$, which is intractable in practice. Results in Table~\ref{tab:ablation study} demonstrate that $Random$ degrades to performance similar to ER, and $Macro$ is also inferior to our proposed estimator. We conjecture that the online CL model concerns more about the distribution within current or short-term input batches than the macro distribution of sequential data stream and memory. Therefore our batch-wise estimator can better exploit the Logit Adjustment technique to improve performance.

\textbf{Hyperparameter sensitivity analysis.} 
We conduct the sensitivity analysis of the hyperparameters $\tau$ and $l$ in our method in Figure~\ref{fig:sensitive}. ER-LAS is robust to a wide range of $l$. In practice, if the distribution fluctuations in the stream can be discerned, we recommend setting short $l$ for streams that change rapidly and vice versa. As to temperature scalar $\tau$, it has distinct impacts on accuracy and forgetting rate. Although a larger $\tau$ can enable models to forget remarkably less, the best accuracy result is achieved around 1.0. Therefore the stability-plasticity trade-off for target applications can be achieved by tuning $\tau$ and $l$ together.
\section{Conclusion}
\label{sec:conclusion}

We discover the class-conditional invariant and prove the optimality of the class-conditional function {that minimizes the class-balanced error} in online class-IL. As a corollary of our theoretical analysis, we introduce Logit Adjusted Softmax with a batch-wise sliding-window estimator to purse the class-conditional function. Extended to online GCL, knowledge of the learned class-conditional function should be preserved for adaptation to domain drift. {Under conditions without model expansion or computationally intensive techniques,} extensive experiments demonstrate that LAS can achieve state-of-the-art performance on various benchmarks with minimal additional computational overhead, confirming the effectiveness and efficiency of our method to mitigate inter-class imbalance. It is effortless to implement LAS and plug it into rehearsal-based methods to correct their recency bias and boost their accuracy. Rehearsal-free approaches with LAS for online CL could be a subject of further study. Furthermore, we will continue to investigate efficient approaches to handling online domain drift, contributing to practical online GCL applications in the real world.
\section{Broader Impacts and Limitations}\label{sec:limitation}

\paragraph{Broader Impacts.} The implication of our research on the community of CL learning is two-fold. Firstly, our proposed LAS method is simple yet highly effective in eliminating the recency bias caused by inter-class imbalance. LAS can be easily implemented and enhance the performance of other online CL methods. Moreover, LAS readily adapts to real-world uses, such as robot environment adaptation, object detection in autonomous driving, and real-time recommendation systems, improving augmented applications’ accuracy. Secondly, we propose partitioning non-stationary data stream learning into the class-conditional and class-prior functions and empirically demonstrate its effectiveness. Our approach lays the framework for future research on online general CL and can provoke more task-free methods to address domain drift. Overall, our work is unlikely to have a negative impact on society.

\paragraph{Limitations.} As the focus of our proposed method in this paper is primarily on addressing the impact of inter-class imbalance and forgetting induced by recency bias on CL performance. We eliminate the imbalanced class-priors to improve performance. However, it can not provide any benefits when faced with domain-IL scenarios with only intra-class domain drift and no inter-class imbalance. Our next research focuses on developing task-free domain-IL methods to address intra-class domain drift efficiently. We hope to integrate our method with task-free domain-IL methods to form a comprehensive solution for online general CL. Another limitation of our approach is that it is not effectively applicable to rehearsal-free online CL. Such a drawback does not stem from the constraints of our theoretical framework but rather from the difficulty in balancing inter-class margins when using the logits adjustment technique to pursue the class-conditional function. For detailed discussions, please refer to the experimental results in \cref{subsec:mnist rehearsal-free}. In the future, we will stick to researching rehearsal-free online CL.

\subsubsection*{Acknowledgments}
The authors would like to thank the anonymous reviewers for their insightful comments.

The research leading to these results has received funding from National Key Research Development Project (2023YFF1104202), National Natural Science Foundation of China (62376155), Shanghai Municipal Science and Technology Research Program (22511105600) and Major Project (2021SHZDZX0102).

\bibliography{main_paper}
\bibliographystyle{tmlr}

\newpage
\appendix
\section{Proof}
\label{sec:proof}

\subsection{Proof in Online Class-incremental Scenarios}

To tackle the issue of inter-class imbalance, extensive research~\citep{Menon2013OnTS, Collell2016RevivingTA, Menon2020LongtailLV} has been conducted on the {optimal classifier that minimizes the class-balanced error} for stable distributions. Actually, previous arts have proposed the following Theorem about {this optimal} classifier:
\begin{theorem}\label{thm:time-invariant bayes}
    For time-invariant distributions, the {optimal estimate that minimizes the class-balanced error} is the class 
    under which the sample probability is most likely:
    \begin{equation}\label{eq:time-invariant bayes}
        \Phi^*\in \mathop{\arg\min}_{\Phi:\mathcal X\rightarrow \mathbb R^{|\mathcal Y|}} \operatorname{CBE}(\Phi,\mathcal Y), \quad \mathop{\arg\max}\limits_{y \in|\mathcal Y|} \Phi_y^*(x)
        =\mathop{\arg\max}\limits_{y \in|\mathcal Y|} \mathbb{P}(x|y)
    \end{equation}
\end{theorem}

Theorem~\ref{thm:time-invariant bayes}~\cite{Menon2013OnTS, Collell2016RevivingTA} states that {this optimal} classifier is independent of arbitrary imbalanced label distributions $\mathbb P(y)$. The class-conditional function in stable distributions naturally minimizes the class-balanced error. From Theorem~\ref{thm:time-invariant bayes} and given the condition of fixed class-conditionals, i.e., $\forall t,\mathbb P(x|y,\rho_t)=\mathbb P(x|y,\rho_0)$, we can derive the proof of Theorem~\ref{thm:bayes optimal} as follows:
\begin{equation}
\begin{split}
    \mathop{\arg\max}\limits_{y \in|\mathcal Y_t|} \Phi_{t,y}^*(x)
    = \mathop{\arg\max}_{y \in|\mathcal Y_t|}\frac{1}{t}\sum_{i=1}^t \mathbb{P}(x|y,\rho_i)
    =\mathop{\arg\max}\limits_{y \in|\mathcal Y_t|} \mathbb{P}(x|y,\rho_t)
\end{split}
\end{equation}

\subsection{Proof in Online General Continual Learning Scenarios}

Without any prior information about the distribution of the test data, we assume that its distribution should conform to a uniformly joint distribution of all observed class distributions. Therefore, the final intra-class distribution is $\frac{1}{t}\sum_{i=1}^t \mathbb{P}(x|y,\rho_i)$. Therefore, the result of Equation~\ref{eq:general bayes optimal} is from definition. 

Let $p_{x|y}$ and $q_{x|y}$ be the underlying distributions the {optimal classifier that minimizes the class-balanced error} and the learned class-conditional function represents, respectively. The class-balanced error gap between the {optimal} classifier $\exp(\Phi_{t,y}^*(x))\propto\mathbb P(x|y,p)=p_{x|y}$ and the learned class-conditional function $\exp(\Phi_{t,y}(x))\propto\mathbb P(x|y,q)=q_{x|y}$ can be formalized as follows: 
\begin{equation}\label{eq:disparity}
\underbrace{|\operatorname{CBE}(\Phi^*,\mathcal Y_t)-\operatorname{CBE}(\Phi,\mathcal Y_t)|}_{\epsilon_t(\Phi^*,\Phi)}\leqslant \underbrace{\frac{1}{|\mathcal Y_t|} \sum\limits_{y \in \mathcal Y_t} \mathbb E_{\rho_t}[\mathbb{E}_{x|y,\rho_t}[\mathop{\arg\max}\limits_{y^{\prime} \in \mathcal Y_t} p_{x|y} \neq \mathop{\arg\max}\limits_{y^{\prime} \in \mathcal Y_t} q_{x|y}]]}_{ d_t(p_{x|y},q_{x|y})}
\end{equation}

Equation~\ref{eq:disparity} describes the disparity $\epsilon_t(\cdot,\cdot)$ from the {optimal} solution by a similarity measure $d_t(\cdot,\cdot)$ in the probability space. Aligning two class-conditionals requires techniques for domain generalization and concept shift. In the future, we will explore efficient class-conditional alignment techniques in the context of online CL.
\section{Algorithm}
\label{sec:algorithm}

We give the algorithm of Experience Replay in Algorithm~\ref{alg:er}. The algorithm of our proposed Logit Adjusted Softmax enhanced Experience Replay in Algorithm~\ref{alg:er las} is mainly based on Algorithm~\ref{alg:er}. We also apply our method to online GCL by combining with knowledge distillation, as shown in Algorithm~\ref{alg:kd las}.
\begin{algorithm}[htb]
   \caption{Experience Replay (ER)~\cite{Chaudhry2019ContinualLW}}
   \label{alg:er}
\begin{algorithmic}
   \STATE {\bfseries Input:} Data stream $\{\mathcal D_t\}_{i=1}^T$
   \STATE {\bfseries Initialize:} Learner $\Phi(\cdot)$, model parameter $\Theta$, memory buffer $\mathcal M_1\leftarrow \{\}$, label set $\mathcal Y_1\leftarrow \{\}$.
   \FOR{$t=1$ {\bfseries to} $T$}
   \STATE {\bfseries Sample incoming batch} $B_t$ {\bfseries from} $\mathcal D_t$
   \STATE $\mathcal Y_t \leftarrow \mathcal Y_{t-1} \cup \text{set}(\{y_i\}_{i=1}^{|B_t|})$
   \STATE $B_t^\mathcal{M}\leftarrow Retrieval(B_t, \mathcal M_t)$
   \STATE $z \leftarrow \Phi(\text{concat}(B_t, B_t^\mathcal{M}),\Theta)$
   \STATE $\text{SGD}( \frac{1}{|B_t|+|B_t^\mathcal{M}|} \sum_{i=1}^{|B_t|+|B_t^\mathcal{M}|}\mathcal L_{\text{CE}}(y_i, z_i), \Theta)$
   \STATE $\mathcal M_{t+1} \leftarrow Update(B_t, \mathcal M_t)$
   \ENDFOR
\end{algorithmic}
\end{algorithm}

\begin{algorithm}[htb]
    \caption{Experience Replay with Logit Adjusted Softmax (ER-LAS)}
    \label{alg:er las}
\begin{algorithmic}
    \STATE {\bfseries Input:} Data stream $\{\mathcal D_t\}_{i=1}^T$, temperature scalar $\tau$, sliding window estimator length $l$
    \STATE {\bfseries Initialize:} Learner $\Phi(\cdot)$, model parameter $\Theta$, memory buffer $\mathcal M_1\leftarrow \{\}$, label set $\mathcal Y_1\leftarrow \{\}$.
    \FOR{$t=1$ {\bfseries to} $T$}
    \STATE {\bfseries Sample incoming batch} $B_t$ {\bfseries from} $\mathcal D_t$
    \STATE $\mathcal Y_t \leftarrow \mathcal Y_{t-1} \cup \text{set}(\{y_i\}_{i=1}^{|B_t|})$ %\COMMENT{// update label set}
    \STATE $B_t^\mathcal{M}\leftarrow Retrieval(B_t, \mathcal M_t)$
    \FOR{$y$ {\bfseries in} $\mathcal Y_t$}
        \STATE $\pi_{y,t} \leftarrow$ {\bfseries compute class-priors from Equation 7}
    \ENDFOR    
    \STATE $z \leftarrow \Phi(\text{concat}(B_t, B_t^\mathcal{M}),\Theta)$
    \STATE $\text{SGD}( \frac{1}{|B_t|+|B_t^\mathcal{M}|} \sum_{i=1}^{|B_t|+|B_t^\mathcal{M}|}\mathcal L_{\text{LAS}}(y_i, z_i), \Theta)$
    \STATE $\mathcal M_{t+1} \leftarrow Update(B_t, \mathcal M_t)$
    \ENDFOR
\end{algorithmic}
\end{algorithm}

\begin{algorithm}[htb]
    \caption{Knowledge distillation with Logit Adjusted Softmax (KD-LAS)}
    \label{alg:kd las}
\begin{algorithmic}
    \STATE {\bfseries Input:} Data stream $\{\mathcal D_t\}_{i=1}^T$, temperature scalar $\tau$, sliding window estimator length $l$
    \STATE {\bfseries Initialize:} Learner $\Phi(\cdot)$, model parameter $\Theta$, memory buffer $\mathcal M_1\leftarrow \{\}$, label set $\mathcal Y_1\leftarrow \{\}$.
    \FOR{$t=1$ {\bfseries to} $T$}
    \STATE {\bfseries Sample incoming batch} $B_t$ {\bfseries from} $\mathcal D_t$
    \STATE $\mathcal Y_t \leftarrow \mathcal Y_{t-1} \cup \text{set}(\{y_i\}_{i=1}^{|B_t|})$ %\COMMENT{// update label set}
    \STATE $B_t^\mathcal{M}\leftarrow Retrieval(B_t, \mathcal M_t)$
    \FOR{$y$ {\bfseries in} $\mathcal Y_t$}
        \STATE $\pi_{y,t} \leftarrow$ {\bfseries compute class-priors from Equation 7}
    \ENDFOR    
    \STATE $z \leftarrow \Phi(\text{concat}(B_t, B_t^\mathcal{M}),\Theta)$
    \STATE $z^{old} \leftarrow \Phi(\text{concat}(B_t, B_t^\mathcal{M}),\Theta^{old})$
    \STATE $\text{SGD}( \frac{1}{|B_t|+|B_t^\mathcal{M}|} \sum_{i=1}^{|B_t|+|B_t^\mathcal{M}|} (\mathcal L_{\text{LAS}}(y_i, z_i)+\mathcal L_{\text{KD}}(z_i, z^{old}_i)), \Theta)$
    \STATE $\mathcal M_{t+1} \leftarrow Update(B_t, \mathcal M_t)$
    \IF {$t$ {\bfseries ends a task.}}
    \STATE {\bfseries Save the old model} $\Theta^{old}\leftarrow\Theta$
    \ENDIF
    \ENDFOR
\end{algorithmic}
\end{algorithm}
\section{Benchmark Details}
\label{sec:benchmark}

\subsection{Dataset Details}

\begin{table}[h]
\caption{Dataset information for CIFAR10, CIFAR100, TinyImageNet, ImageNet, and iNaturalist.}
\label{tab:dataset}
\begin{center}
\begin{tabular}{l|cccc}
\toprule
\textbf{Dataset}      & Image Size & \# Train & \# Test & \# Class \\
\midrule
CIFAR10~\cite{Krizhevsky2009LearningML}      & $3\times32\times32$    & 50,000 & 10,000 & 10\\
CIFAR100~\cite{Krizhevsky2009LearningML}     & $3\times32\times32$    & 50,000 & 10,000 & 100\\
TinyImageNet~\cite{Le2015TinyIV} & $3\times64\times64$    & 100,000 & 10,000 & 200\\
ImageNet~\cite{Deng2009ImageNetAL}     & $3\times224\times224$  & 1,281,167 & 50,000 & 1000 \\
iNaturalist~\cite{Horn2017TheIS}  & $3\times299\times299$  & 579,184   & 95,986 & 5089\\
\bottomrule
\end{tabular}
\end{center}
\end{table}

We list the image size, the total number of training samples, the total number of test samples, and the total number of classes for the 5 datasets (CIFAR10~\cite{Krizhevsky2009LearningML}, CIFAR100~\cite{Krizhevsky2009LearningML}, TinyImageNet~\cite{Le2015TinyIV}, ImageNet~\cite{Deng2009ImageNetAL}, and iNaturalist~\cite{Horn2017TheIS}) in Table~\ref{tab:dataset}. In the former four class-balanced datasets, each category contains an equivalent number of training and test samples. However, within iNaturalist, an inherent imbalance exists between classes, posing a greater challenge. We download the dataset of iNaturalist from \url{https://github.com/visipedia/inat_comp}. 

\subsection{Continual Learning Setup Details}

\begin{figure}[htb]
    \centering 
    \includegraphics[width=0.8\linewidth]{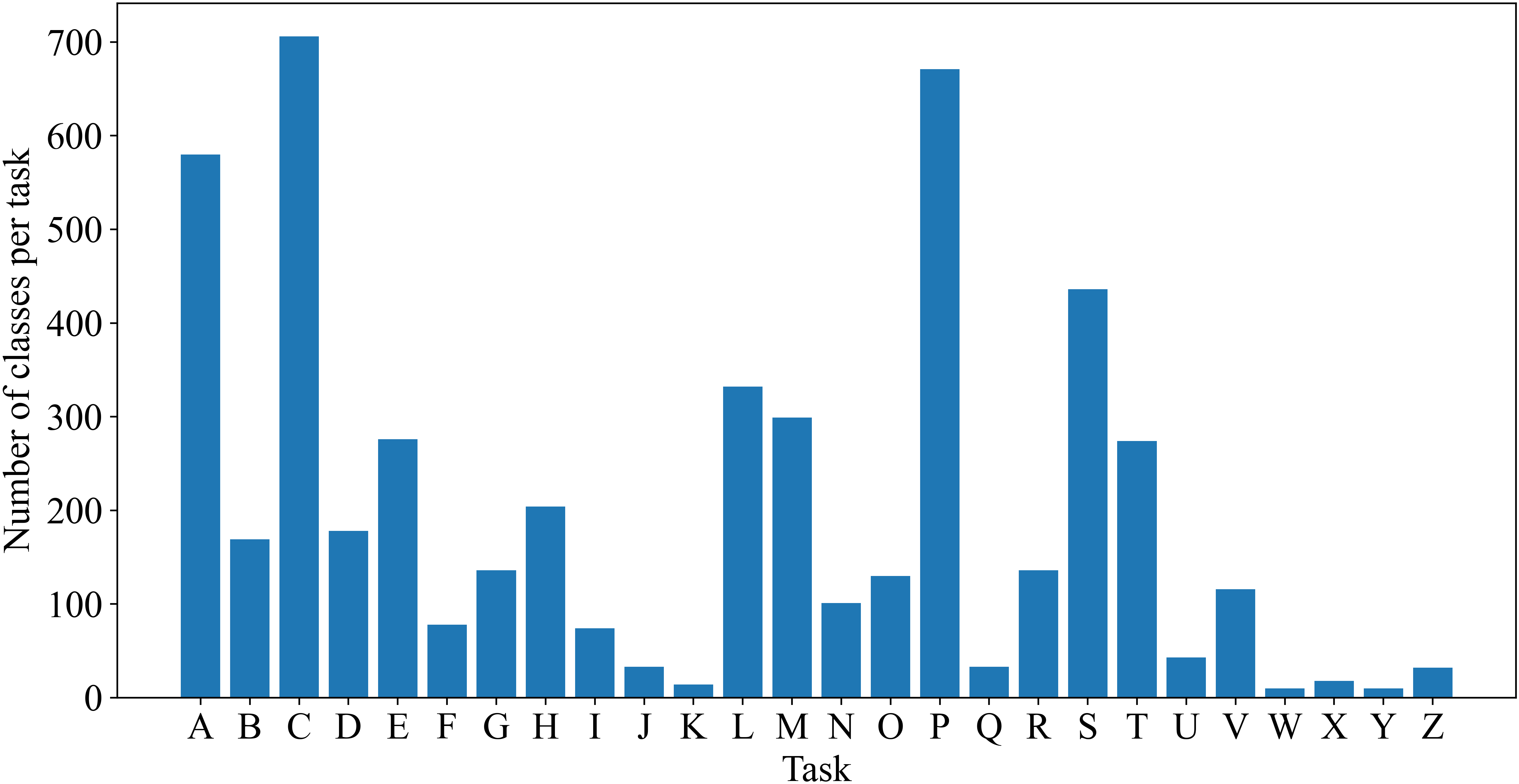}
    \caption{The number of classes per task in divided iNaturalist.
    Each one of these 26 tasks contains categories with the same corresponding initial letter.
    }
    \label{fig:inat class}
\end{figure}

In \textbf{online class-IL}~\cite{Aljundi2019OnlineCL}, classes of CIFAR10, CIFAR100, TinyImageNet, and ImageNet are evenly split from the total into each task. And the classes in different tasks are disjoint. For example, C-CIFAR10 (5 tasks) is split into 5 disjoint tasks with 2 classes each. As a result, the numbers of training samples, testing samples, and classes are the same in each task, except iNaturalist. We divide the iNaturalist into 26 disjoint tasks according to the initial letter of the category. The numbers of classes in each task are shown in Figure~\ref{fig:inat class}. It shows that the number of classes varies significantly among each task. Noting that the classes within each task are also imbalanced. The comprehensive inter-class imbalance issues of C-iNaturalist (26 tasks) pose great challenges to online CL methods.

In \textbf{online blurry CL}~\cite{Koh2021OnlineCL}, the classes are divided into $N_{\text{blurry}}\%$ disjoint part and $(100-N_{\text{blurry}}\%)$ blurry part. The classes that belong to the disjoint part will only appear in fixed tasks, while all other classes in the blurry part will occur throughout the data stream. In each task, $(\text{\#train}-(T-1)*M_{\text{blurry}})$ instances will be sampled from the training data of head blurry classes and $M_{\text{blurry}}$ instances will be sampled from the training data of remaining blurry classes, which forms the apparently class-imbalanced blurry part samples. The classes in blurry part play the role of head classes in turn across different tasks. During inference, the model will predict on test samples from all currently observed classes. We split CIFAR100 and TinyImageNet into 10 blurry tasks according to \cite{Koh2021OnlineCL} with fixed disjoint ratio $N_{\text{blurry}}=50$ and blurry level $M_{\text{blurry}}=10$. Next we take B-CIFAR100 (10 tasks) as an example.

In B-CIFAR100 (10 tasks, $N_{\text{blurry}}=50$, $M_{\text{blurry}}=10$, $\text{\#train}=500$ per class, $\text{\#class}=100$), the disjoint part contains 50 classes, and each task possesses 5 disjoint classes of training data. On the other hand, the blurry part comprises the other 50 classes, and each task has 5 head classes. The head classes contain $500-9*10=410$ training samples, whereas the remaining 45 blurry classes only have 10 training samples each for the current task. Therefore the model in this setup will continuously observes disjoint new classes as stream flows and imbalanced classes overlap across all tasks, encountering a severe problem of inter-class imbalance.

\begin{figure}[htb]
    \centering 
    \includegraphics[width=0.99\linewidth]{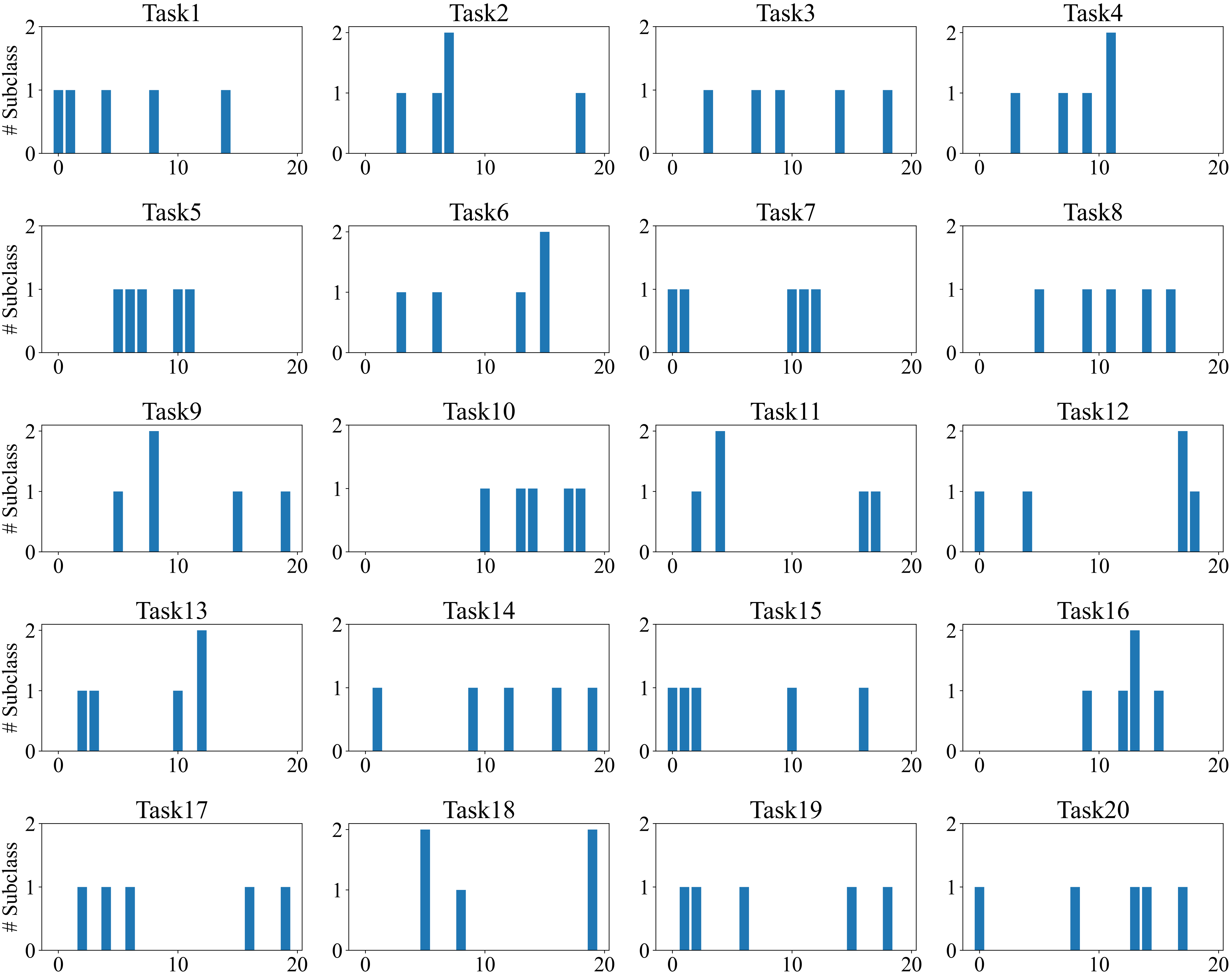}
    \caption{An illustration of the occurrence of subclasses within each superclass for every task in {S}-CIFAR100 (20 tasks). The $y$-axis represents the number of occurrences of subclasses. The $x$-axis represents the 20 superclasses. Worth noting that each subclass is a distinct domain.
    }
    \label{fig:general}
\end{figure}

In \textbf{online {Sum-Class-Domain} CL}~\cite{Xie2022GeneralIL}, incoming data contains images from new classes and new domains. We only apply online {Sum-Class-Domain} CL on CIFAR100. Similar to the online class-IL setup, we partition the CIFAR100 dataset into 20 tasks, each with 5 subclasses. However, the model is required to predict superclasses, with each subclass representing a distinct domain within them. Each domain of superclass has the same number of training samples. As depicted in Figure~\ref{fig:general}, different superclasses appear in various tasks. Also, varying number of superclasses occur in each tasks. And the distribution within each superclass changes across different domains. Therefore, {S}-CIFAR100 (20 tasks) possesses both inter-class imbalance and intra-class domain drift, i.e, changing class-priors and class-conditionals. {Next, we discuss scenarios where mixed class and domain distributions jointly change, presenting a greater challenge than the sum of individual online class- and domain-IL learning problems. We refer to this case as "Mix-Class-Domain." This increased difficulty arises due to the possible coupling between domain shift and class distribution shift in Mix-Class-Domain CL.
\begin{itemize}
    \item Class-IL: The joint distribution of sample $x$ and label $y$ can be written as $p (x, y, t) = p (x | y, t), p (y, t)$, where $p(x|y,t)$ is constant throughout time $t$. $p(y,t)$ changes with data flow and brings about inter-class imbalance.
    \item Domain-IL: A common assumption in domain-IL is that each class has the latent domain indicator $z$. The joint distribution of sample $x$, label $y$, and domain indicator $z$ can be decomposed into $p (x, y, z, t) = p (x | z, y, t), p (z | y, t), p (y, t)$, where only $p (z | y, t)$ changes over time $t$, while $p (x | z, y, t)$ and $p (y, t)$ keep constant.
    \item Sum-Class-Domain: This is a special case of the Mix-Class-Domain CL problems, where domain shift and class distribution shift are separable. In the decomposition of the joint distribution, $p (x, y, z, t) = p (x | z, y, t), p (z | y, t), p (y, t)$, only $p(x|z,y,t)$ remains invariant and independent of $t$, while $p(z|y,t)$ and $p(y,t)$ vary over time, giving rise to intra-class imbalance and inter-class imbalance issues, respectively.
    \item Mix-Class-Domain: We consider the inherent coupling between domain indicators $z$ and class labels $y$, rather than their mutual independence which allows for decomposition. In this case, the joint distribution can only be decomposed as $p(x,y,z,t)=p(x|z,y,t)p(z,y,t)$. The fact that domain indicators and class labels are inseparable forms a more challenging problem than the separable Sum-Class-Domain CL, which may lead to more forgetting.
\end{itemize}}
{Unfortunately, there is also a lack of standard benchmarks for the indecomposable case.} As one of the most challenging and realistic scenarios for practical applications, it deserves further in-depth investigation in future research.

\section{Metrics Details}
\label{sec:metrics}

Assume test samples of task $j$ is {$S_j=\left\{(x_n, y_n)\right\}_{n=1}^{N^j}$}.
% $S_j=\{x_n,y_n\}_{n=1}^N$. 
The number of test samples in each class $y$ {for task $j$ is $N_y^j$}. The model trained on task $i$ is $\Phi_i$. The seen class set at task $i$ is $\mathcal Y_i$. The accuracy $a_{i,j}$ on task $j$ after training on task $i$ is formalized as follows:
% \begin{equation}\label{eq:a_i_j}
% a_{i,j}
% = \frac{1}{N}\sum_{n=1}^N
% \mathbbm{1}(\mathop{\arg\max}\limits_{y^{\prime}\in\mathcal Y_i}\Phi_{i,y^{\prime}}(x_n)=y_n).
% \end{equation}
\begin{equation}\label{eq:a_i_j}
a_{i, j}=\frac{1}{N^j} \sum_{S_j} \mathbbm{1}\left(\underset{y^{\prime} \in \mathcal{Y}_i}{\arg \max } \Phi_{i, y^{\prime}}\left(x_n\right)=y_n\right).
\end{equation}
For long-tailed data streams with inherent inter-class imbalance, we also consider a more appropriate metric, namely class-balanced accuracy $a_{i,j}^{\text{cbl}}$, instead of standard accuracy $a_{i,j}$ for evaluation. Class-balanced accuracy excludes prior class imbalances and prevents the overestimation of trivial solutions with high probabilities for major classes. 
% \begin{equation}\label{eq:a_i_j class balanced}
% a_{i,j}^{\text{cbl}} 
% = \frac{1}{|\mathcal Y_i|}\sum_{y\in \mathcal Y_i}\frac{1}{N_y}\sum_{\{(x_n,y_n)|y_n=y\}} 
% \mathbbm{1}(\mathop{\arg\max}\limits_{y^{\prime}\in\mathcal Y_i}\Phi_{i,y^{\prime}}(x_n)=y)
% .\end{equation}
\begin{equation}\label{eq:a_i_j class balanced}
a_{i, j}^{\mathrm{cbl}}=\frac{1}{\left|\mathcal{Y}_i\right|} \sum_{y \in \mathcal{Y}_i} \frac{1}{N^j_y} \sum_{\left\{(x_n, y_n) \mid(x_n,y_n)\in S_j, y_n=y\right\}} \mathbbm{1}\left(\underset{y^{\prime} \in \mathcal{Y}_i}{\arg \max } \Phi_{i, y^{\prime}}\left(x_n\right)=y\right)
.\end{equation}
The corresponding final average accuracy $A_T$ and final average class-balanced accuracy $A_T^{\text{cbl}}$ can be calculated as follows: 
\begin{equation}\label{eq:A_T}
    A_T = \frac{1}{T} \sum_ {j=1}^T a_{T,j},
\end{equation}
\begin{equation}\label{eq:A_T cbl}
    A_T^{\text{cbl}} = \frac{1}{T} \sum_ {j=1}^T a_{T,j}^{\text{cbl}}.
\end{equation}
{In datasets such as CIFAR-10 and CIFAR-100, where class-priors are uniform, the final average accuracy is equal to the final average class-balanced accuracy, consistent with our analysis.}

The final average forgetting $F_T$ can be computed~\cite{Chaudhry2020ContinualLI} as follows:
\begin{equation}\label{eq:F_t}
    F_T=\frac{1}{T-1} \sum_{j=1}^{T-1} \max_{i \in\{1, \dots, T-1\}}\left(a_{i,j}-a_{T,j}\right).
\end{equation} 

We follow \cite{Koh2021OnlineCL} to add the Area Under the Curve of Accuracy $A_{\text{AUC}}$ in the online blurry CL setup. $A_{\text{AUC}}$ is the average accuracy to \{\# of samples\}. We simplify the calculation of $A_{\text{AUC}}$ by replacing \{\# of samples\} with \{\# of steps\}. Then this metric can be calculated as follows:
\begin{equation}\label{eq:A_AUC}
    A_{\text{AUC}}=\frac{1}{N}\sum_{k=1}^Kf(k\cdot\Delta n)\cdot \Delta n,
\end{equation}
where $N$ represents the total number of training steps, $\Delta n$ denotes that we sample the accuracy $f(\cdot)$ of the model every $n$ steps, and $K$ is the total number of sample intervals. We set $\Delta n =5$ in the experiments.
\section{Implementation Details}
\label{sec:implementation}

\subsection{Baseline Implementation}

We as follows list the hyperparameter configurations for the baseline methods mentioned in this paper, along with their sources of code implementation.

For ER~\citep{Chaudhry2019ContinualLW}, we set the learning rate as $0.03$. The code source is \url{https://github.com/aimagelab/mammoth}.

For DER++~\citep{Buzzega2020DarkEF}, we set the learning rate as $0.03$. $\alpha$ is set to $0.1$, and $\beta$ is set to $0.5$. The code source is \url{https://github.com/aimagelab/mammoth}.

For MRO~\citep{chrysakis2023online}, we set the learning rate as $0.03$. The code source is \url{https://github.com/aimagelab/mammoth}.

For SS-IL~\citep{Ahn2020SSILSS}, we set the learning rate as $0.03$. We update the teacher model every $100$ steps. The code source is \url{https://github.com/hongjoon0805/SS-IL-Official}.

For CLIB~\citep{Koh2021OnlineCL}, we set the learning rate as $0.03$. The period between sample-wise importance updates is set to $3$. The code source is \url{https://github.com/naver-ai/i-Blurry}.

For ER-ACE~\citep{Caccia2021NewIO}, we set the learning rate as $0.03$. The code source is \url{https://github.com/pclucas14/AML}.

For ER-OBC~\citep{chrysakis2023online}, we set the learning rate as $0.03$ for both training and bias correction. The code source is \url{https://github.com/chrysakis/OBC}.

{For ER-CBA~\citep{wang2023cba}, we set the learning rate as $0.001$ for C-CIFAR10, and 0.01 for C-CIFAR100 and C-TinyImageNet. The code source is \url{https://github.com/wqza/CBA-online-CL}.}

For MIR~\citep{Aljundi2019OnlineCL}, we set the learning rate as $0.03$. The number of subsampling in replay is $160$. The code source is \url{https://github.com/optimass/Maximally_Interfered_Retrieval}.

For ASER$_\mu$~\citep{Shim2020OnlineCC}, we set the learning rate as $0.03$. The number of nearest neighbors $K$ to perform ASER is $5$. We use mean values of adversarial Shapley values and cooperative Shapley values. The maximum number of samples per class for random sampling is $6.0$ times of incoming batch size. The code source is \url{https://github.com/RaptorMai/online-continual-learning}.

For OCS~\citep{yoon2022online}, we set the learning rate as $0.03$. The hyperparameter $\tau$ that controls the degree of model plasticity and stability is set to $1000.0$. The code source is \url{https://openreview.net/forum?id=f9D-5WNG4Nv}.

For LwF~\citep{Li2016LearningWF}, we set the learning rate as $0.03$. The penalty weight $\alpha$ is set to $0.5$ and the temperature scalar is set to $2.0$. The code source is \url{https://github.com/aimagelab/mammoth}.

For LURIC~\citep{Hou2019LearningAU}, we set the learning rate as $0.03$. $\lambda_{\text{base}}$ is set to $5.0$ for all the experiments. The code source is \url{https://github.com/hshustc/CVPR19_Incremental_Learning}.

For GeoDL~\citep{Simon2021OnLT}, we set the learning rate as $0.03$. The adaptive weight $\beta$ is set to $5.0$. The code source is \url{https://github.com/chrysts/geodesic_continual_learning}.

We also list the hyperparameter configuration for the baseline methods used in this appendix with their sources of code implementation.

For SCR~\citep{Mai2021SupervisedCR}, we set the learning rate as $0.03$ and the temperature as $\tau=0.07$. The code source is \url{https://github.com/RaptorMai/online-continual-learning}.

For OCM~\citep{Guo2022OnlineCL}, we use Adam optimizer and set the learning rate as $0.001$. The code source is \url{https://github.com/gydpku/OCM}.

For BiC~\citep{Wu2019LargeSI}, we set the learning rate as $0.03$. We split $10\%$ of the training data into a validation set for training the bias injector with 50 epochs. The softmax temperature $T$ is $2.0$. Distillation loss is also applied after bias correction. The code source is \url{https://github.com/sairin1202/BIC}.

For E2E\citep{Castro2018EndtoEndIL}, we set the learning rate as 0.03. In the process of balanced fine-tuning, we set the learning rate as 0.003 and train 30 epochs. The code source is \url{https://github.com/PatrickZH/End-to-End-Incremental-Learning}.

For IL2M\citep{Belouadah2019IL2MCI}, we set the learning rate as 0.03. We calculate the mean and variance of each batch online to re-scale the outputs. The code source is \url{https://github.com/EdenBelouadah/class-incremental-learning}.

For LUCIR~\citep{Hou2019LearningAU}, we set the learning rate as $0.03$. $\lambda_{\text{base}}$ is set to $5.0$, $K$ is set to $2$, and $m$ is set to $0.5$ for all the experiments. The code source is \url{https://github.com/hshustc/CVPR19_Incremental_Learning}.

\subsection{Ablation Implementation}

In the ablation study of \S\ref{subsec:ablation studies}, we employ two extreme variants of LAS, along with two estimators. $Random$ estimator randomly assigns class-priors. $Macro$ uses statistical information to assign class-priors. Now, we elaborate on how these four methods are implemented. Recalling our proposed Logit Adjusted Softmax cross-entropy loss in Equation~\ref{eq:logit adjusted softmax ce}. 
\begin{equation}\label{eq:logit adjusted softmax ce 2}
    \mathcal L_{\text{LAS}}(y, \Phi(x))
    =-\log \frac{e^{\Phi_y(x)+\tau \cdot \log \pi_{y,t}}}{\sum_{y^{\prime} \in\mathcal Y_t} e^{\Phi_{y^{\prime}}(x)+\tau \cdot \log \pi_{y^{\prime},t}}}
    =\log [1+\sum_{y^{\prime} \neq y}\left(\frac{\pi_{y^{\prime},t}}{\pi_{y,t}}\right)^\tau \cdot e^{\left(\Phi_{y^{\prime}}(x)-\Phi_y(x)\right)}].
\end{equation}

\paragraph{$\tau=0$} is a simple special case that sets the temperature scalar to $0$.

\paragraph{$\tau=\infty$} needs modification because directly setting the hyperparameter $\tau$ to a large value to pursue $\infty$ would cause troubles when ${\pi_{y^{\prime},t}}/{\pi_{y,t}}>1$, as it would lead to an infinity coefficient and result in gradient explosion, obstructing the gradient descent optimization algorithm. Therefore, as shown in Equation~\ref{eq:las coef}, we set the coefficient to $0$ only when ${\pi_{y^{\prime},t}}/{\pi_{y,t}}<1$, while retaining $\tau=1$ for all other situations to enable successful model training. The significantly low forgetting rate and competitive accuracy observed in the experimental results suggest that this approach closely approximates $\tau=\infty$ as expected.
\begin{equation}\label{eq:las coef}
    \left(\pi_{y^{\prime},t}/\pi_{y,t}\right)^\tau = 
    \begin{cases}
    0, &\left(\pi_{y^{\prime},t}/\pi_{y,t}\right) < 1\\
    \left(\pi_{y^{\prime},t}/\pi_{y,t}\right), &\text{otherwise}
    \end{cases}.
\end{equation}

\paragraph{$Random$} samples each prior of seen classes from a uniform distribution $U[0,1]$. Then they are normalized to $1$.

\paragraph{$Macro$} computes the joint label distribution by taking into account the occurrence frequencies of each class in the current data stream, as well as the label probabilities in the memory buffer, to serve as the current class-priors. It is worth noting that since the distribution of the data stream is unknown during training, $Macro$ cannot be directly obtained and serves only as a reference for comparing and validating the necessity of batch-wise estimators. For instance, in C-CIFAR (5 tasks), when it comes to the 2$^\text{nd}$ task, 2 classes in the data stream are of the same quantity, and the 2 classes in the memory buffer also contain a similar number of samples from the previous task. The incoming and buffer batch sizes are also the same. At this point, the 4 classes probabilities that may appear in the input batch are all equally likely, i.e., $1/4$. When it comes to the 5$^\text{th}$ task, the data stream still consists of 2 classes with the same label probabilities, but the memory buffer now stores 8 classes that have appeared before. Therefore, it can be calculated that the class-priors of the 2 classes in the data stream are $1/4$, while the class-priors of the 8 classes in the memory buffer are $1/16$. $Macro$ represents a statistical oracle, but experiments show that its performance is inferior to batch-wise estimators, indicating that in online CL, the model may pay more attention to the label distributions within each batch rather than the label distributions across the sequential tasks.
\section{More Experimental Results}
\label{sec:more experiment}

\subsection{Results on C-MNIST}
\label{subsec:mnist class-il}
We provide the results for MNIST of the online class-IL setting in Table~\ref{tab:class cl mnist}. ER-LAS still achieves competitive performance, highlighting the effectiveness of our method in addressing simple online CL problems. However, we observe that on small datasets like MNIST, the performance improvement achieved by our method is limited. This limitation arises because the tasks in MNIST are relatively easy, and the forgetting caused by inter-class imbalance is not prominent. Therefore, in our experiments, we have primarily focused on more challenging scenarios with considerable classes or large-scale datasets like ImageNet. Considering that online CL aims to handle potentially infinite data streams, we believe that scalability to large-scale datasets is crucial in validating the effectiveness of online CL algorithms. 

{
\begin{table*}[htb]
    \caption{{Final average accuracy $A_T$ (higher is better) on C-MNIST (5 tasks). $M$ is memory size.}}
    \label{tab:class cl mnist}
    \small
    \centering
    \begin{tabular}{@{}l | ccc@{}} 
    \toprule
    \textbf{Method}          & $M=0.5k$ & $M=1k$ & $M=2k$ \\
    % \midrule
    \midrule
    ER              &$87.0$\tiny{$\pm0.2$}&$88.0$\tiny{$\pm0.2$}&$90.7$\tiny{$\pm0.1$}\\
    DER++           &$\mathbf{92.3}$\tiny{$\mathbf{\pm0.1}$}&$\mathbf{93.9}$\tiny{$\mathbf{\pm0.0}$}&$\mathbf{94.2}$\tiny{$\mathbf{\pm0.1}$}\\
    MRO             &$87.4$\tiny{$\pm0.2$}&$88.9$\tiny{$\pm0.1$}&$92.6$\tiny{$\pm0.1$}\\
    SS-IL           &$88.7$\tiny{$\pm0.3$}&$90.3$\tiny{$\pm0.2$}&$91.7$\tiny{$\pm0.1$}\\
    CLIB            &$88.4$\tiny{$\pm0.3$}&$90.6$\tiny{$\pm0.0$}&$91.9$\tiny{$\pm0.0$}\\
    ER-ACE          &$90.4$\tiny{$\pm0.0$}&$92.4$\tiny{$\pm0.1$}&$93.8$\tiny{$\pm0.2$}\\
    ER-OBC          &$90.0$\tiny{$\pm0.1$}&$89.7$\tiny{$\pm0.1$}&$91.6$\tiny{$\pm0.4$}\\
    ER-CBA          &$90.1$\tiny{$\pm0.3$}&$90.2$\tiny{$\pm0.1$}&$91.3$\tiny{$\pm0.1$}\\
    \rowcolor{myblue}
    ER-LAS          &$91.7$\tiny{$\pm0.1$}&$92.8$\tiny{$\pm0.1$}&$94.0$\tiny{$\pm0.1$}\\
    \bottomrule
    \end{tabular}
\end{table*}
}

\subsection{Results on MNIST without Rehearsal}
\label{subsec:mnist rehearsal-free}
Despite the current mainstream online continual learning methods incorporating memory replay of samples and achieving satisfactory performance, a range of studies~\citep{zajkac2023prediction,Li2016LearningWF,Kirkpatrick2016OvercomingCF,zeno2018task} have also explored effective continual learning approaches without rehearsal. Our theoretical foundation of the class-conditional function paves the way for our rehearsal-free applications. In our proposed LAS loss function, the use of replayed samples is not necessary, allowing for direct application in online continual learning scenarios without replay. We conducted experiments on the C-MNIST dataset with no memory. The results in Table C show that LAS outperforms previously proposed methods. PEC achieves superior performance to LAS, benefitting from the expansion of new models for each class. Although our theorems hold without the need for rehearsal, the implementation of the logit adjustment technique requires support from replay data. As indicated on the right-hand side of \cref{eq:logit adjusted softmax ce} in the original paper, LAS adjusts inter-class classification margins by leveraging the imbalanced class-priors between major and minor classes, achieving balanced learning. When the number of minor classes decreases to zero, LAS fails to adjust the inter-class classification margins and degrades into learning based only on the currently encountered classes.

{
\begin{table*}[htb]
    \caption{{Final average accuracy $A_T$ (higher is better) on C-MNIST (5 tasks) without rehearsal.}}
    \label{tab:rehearsal-free mnist}
    \small
    \centering
    \begin{tabular}{@{}l | c@{}} 
    \toprule
    \textbf{Method}          & $A_T \uparrow$ \\
    % \midrule
    \midrule
    LwF             &$19.8$\tiny{$\pm0.0$}\\
    EWC             &$19.8$\tiny{$\pm0.0$}\\
    Labels trick    &$45.7$\tiny{$\pm3.5$}\\
    PEC             &$\mathbf{92.3}$\tiny{$\mathbf{\pm0.1}$}\\
    \rowcolor{myblue}
    LAS             &$48.4$\tiny{$\pm1.2$}\\
    \bottomrule
    \end{tabular}
\end{table*}
}

\subsection{Results when Batch Sizes are varied}
\label{subsec:more batch}

While typically samples arrive one by one in the online learning data stream, advanced algorithms\citep{Caccia2021NewIO,chrysakis2023online} are commonly designed to update the model by accumulating a certain number of incoming samples as a batch. This is because per-batch updating is generally more advantageous for model optimization convergence and well-defined classification boundaries than updating on each individual sample. However, in some situations with constrained computational resources, only very small batch sizes are available or the batch sizes vary. Based on this concern, we consider two setups related to changing the batch size: one is various batch sizes for the entire online training process, and the other is varying the batch size during training. We conduct experiments on online C-CIFAR10. We begin with brief introductions to these two setups.
\begin{enumerate}
    \item Evaluating the batch size change for the entire online training process examines the macro robustness of our method to the hyperparameter of batch size. In the manuscript, we set both incoming and buffer batch sizes to 32. We now experiment with corresponding batch sizes of 4, 16, 64. Smaller batch sizes bring more gradient updates for the model, but each contains less information for forming inter-class margins. Larger batch sizes may lead to overfitting on the memory buffer, thereby reducing performance.
    \item Varying the batch size throughout the entire online training process examines the micro robustness of the batch size. This is a practical scenario where the frequency of incoming data may vary at different stages, requiring time-varying batch sizes. In this experiment, we only vary incoming batch sizes while keeping buffer batch sizes at 32. We consider 3 cases of changing incoming batch sizes:
    \begin{itemize}
        \item Increasing incoming batch sizes during training, specifically for C-CIFAR10: 2, 4, 8, 16, 32, as \textbf{Increase}. The inter-class imbalance issue intensifies.
        \item Decreasing incoming batch sizes during training, specifically for C-CIFAR10: 32, 16, 8, 4, 2, as \textbf{Decrease}. The inter-class imbalance issue is alleviated.
        \item Randomly sampling incoming batch sizes from a uniform distribution $U[2,32]$ at each stage, as \textbf{Random}. This is a fusion of the previous two cases, where the impact of inter-class imbalance varies during training.
    \end{itemize}
\end{enumerate}

\begin{table*}[htb]
    \caption{Comparison of final average accuracy on online C-CIFAR10 with various batch sizes. In the manuscript, we set both incoming and buffer batch sizes to 32. We experiment with corresponding batch sizes of 4, 16, and 64. Experimental settings are the same as in Table~\ref{tab:class cl}. Memory sizes are $M=1k$.}
    \label{tab:various bs}
    \centering
    \begin{tabular}{@{}l | c | c | c | c@{}} 
    \toprule
    \textbf{Batch size}         &4 &16 & 32 & 64\\
    \midrule
        ER &$54.0\pm1.8$ &$52.4\pm1.8$ &$45.4\pm1.8$  &$45.4\pm1.9$\\
        ER-ACE &$55.1\pm1.9$ &$56.7\pm2.1$ &$48.1\pm1.1$  &$46.2\pm1.9$\\
        ER-OBC &$48.6\pm1.8$ &$54.7\pm1.4$ &$46.4\pm0.6$  &$39.0\pm1.8$\\
        \rowcolor{myblue}
        ER-LAS&$\mathbf{59.2\pm1.2}$ &$\mathbf{57.5\pm1.3}$ &$\mathbf{55.3\pm1.6}$  &$\mathbf{53.1\pm1.2}$\\
    \bottomrule
    \end{tabular}
\end{table*}

\begin{table*}[htb]
    \caption{Comparison of final average accuracy on online C-CIFAR10 with varying batch sizes during training. We only vary incoming batch sizes while keeping buffer batch sizes as 32: \textbf{Increase} incoming batch sizes during training, i.e., 2, 4, 8, 16, 32. \textbf{Decrease} incoming batch sizes during training, i.e., 32, 16, 8, 4, 2. \textbf{Random}ly sampling incoming batch sizes from a uniform distribution $U[2, 32]$ at each stage. Experimental settings are the same as in Table~\ref{tab:class cl}. Memory sizes are $M=1k$.}
    \label{tab:varying bs}
    \centering
    \begin{tabular}{@{}l | c | c | c@{}} 
    \toprule
    \textbf{Incoming batch size}         &Increase &Decrease &Random\\
    \midrule
        ER &$52.7\pm1.8$ &$65.2\pm1.9$  &$55.1\pm1.8$\\
        ER-ACE &$50.8\pm1.2$ &$62.5\pm1.1$  &$55.1\pm1.9$\\
        ER-OBC &$53.1\pm1.4$ &$65.3\pm1.1$  &$51.4\pm1.7$\\
        \rowcolor{myblue}
        ER-LAS&$\mathbf{59.8\pm1.1}$ &$\mathbf{65.7\pm0.7}$  &$\mathbf{59.3\pm1.9}$\\
    \bottomrule
    \end{tabular}
\end{table*}

The results in Table~\ref{tab:various bs} and Table~\ref{tab:varying bs} show that ER-LAS consistently achieves the best accuracy across various and varying batch sizes, highlighting the robustness of our method to batch size variations. In theory, changing batch sizes or the variation of batch sizes during training poses no threat to our principle of mitigating inter-class imbalance through the elimination of class-priors. It only affects our estimation of time-varying class-priors. However, the ablation study of estimators in \S\ref{subsec:ablation studies} indicates that online CL models may pay more attention to the current input class distributions. Therefore, our designed batch-wise estimator can timely provide effective approximation at various batch sizes. The potential issues may lie in the cases where batch sizes become extremely small, such as 1. Following, we discuss this problem in detail and provide recommendations for improvement.

In fact, training on a single incoming sample goes against the theory of traditional stochastic gradient descent, which may harm model convergence and hinder the establishment of classification boundaries. Therefore, we maintain the incoming batch size of 1 and consider concatenating various numbers of buffer batch sizes to ensure valid training and practical performance. The experiments are conducted on online C-CIFAR10 in order to explore the impact of changed buffer batch sizes on a single incoming batch size.

\begin{table*}[htb]
    \caption{Comparison of final average accuracy on online C-CIFAR10 with fixed incoming batch sizes of 1 and various buffer batch sizes. Experimental settings are the same as in Table~\ref{tab:class cl}. Memory sizes are $M=1k$.}
    \label{tab:one incoming}
    \centering
    \begin{tabular}{@{}l | c | c | c | c@{}} 
    \toprule
    \textbf{Buffer batch size}         & 1 &4 &16 & 64\\
    \midrule
        ER &$\mathbf{39.3\pm2.0}$ &$62.4\pm2.0$ &$63.7\pm1.8$  &$60.6\pm1.9$\\
        ER-ACE &$27.9\pm2.1$ &$57.7\pm1.7$ &$58.1\pm1.8$  &$54.5\pm1.7$\\
        ER-OBC &$33.2\pm1.9$ &$64.3\pm1.8$ &$65.3\pm1.8$  &$60.9\pm1.7$\\
        \rowcolor{myblue}
        ER-LAS&$36.9\pm1.5$ &$\mathbf{66.4\pm1.5}$ &$\mathbf{67.2\pm1.5}$  &$\mathbf{62.2\pm1.3}$\\
    \bottomrule
    \end{tabular}
\end{table*}

The results in Table~\ref{tab:one incoming} show that when both incoming and buffer batch sizes are 1, ER-LAS performs slightly worse than the ER baseline. Nevertheless, simply increasing buffer batch sizes can enable ER-LAS to achieve the highest accuracy. This is because the case of extremely small batch sizes of 1 affects our estimation of time-varying class-priors and hinders the construction of classification margins, where slightly increasing buffer batch sizes can serve as an effective approach to refresh our method. Noting that excessive buffer batch sizes can lead to overfitting on the memory buffer and harm performance, as shown in the rightmost column of Table ~\ref{tab:one incoming}.

\subsection{Comparison to Inter-class Imbalance Mitigation Methods}
\label{subsec:more mitigation}

We mentioned the differences between LAS and other class imbalance mitigation methods from an analytical perspective in \S\ref{sec:related work}. We worry that since these methods have not been deliberately designed and applied to online CL in previous works, our direct application may lack credibility and endorsement. As a result, we do not compare with these methods in experiments. However, we have indeed conducted experiments with them in the preliminary exploration phase of our method. Here, we briefly describe our applications and provide experimental comparisons and analysis. We compare four class imbalance mitigation methods originally for stable distributions. We refer to \cite{Cui2019ClassBalancedLB} and apply the Class-Balanced loss, which re-weights the loss terms of each class based on the input class distribution, as \textbf{ER-CBL} of Loss weighting. We refer to \cite{Kang2019DecouplingRA} and normalize the weights of classifiers with $\lVert w_y\rVert_2$, as \textbf{ER-WN} of Weight normalization. We perform upsampling on the buffer samples and downsampling by randomly ignoring some incoming samples to maintain consistent input class distributions, as \textbf{ER-Up} and \textbf{ER-Down} of Resampling~\citep{Kubt1997AddressingTC}.

\begin{table*}[htb]
    \caption{Comparison of final average accuracy by ER, ER-LAS, and imbalance mitigation methods. Experimental settings are the same as in Table~\ref{tab:class cl}. Memory sizes are $M=1k$.}
    \label{tab:imbalance mitigation}
    \centering
    \begin{tabular}{@{}l | c | c | c@{}} 
    \toprule
    \textbf{Dataset}    &C-CIFAR10&C-CIFAR100&C-TinyImageNet\\
    \midrule
        ER &$45.4\pm1.8$ &$16.5\pm0.4$  &$11.0\pm0.2$\\
        ER-CBL &$48.1\pm1.6$ &$18.8\pm0.2$  &$11.1\pm0.2$\\
        ER-WN &$46.1\pm1.5$ &$16.6\pm0.8$  &$11.0\pm0.1$\\
        ER-Up &$53.6\pm1.6$ &$23.4\pm0.5$  &$15.0\pm0.1$\\
        ER-Down &$48.2\pm1.6$ &$18.9\pm0.3$  &$13.4\pm0.2$\\
        \rowcolor{myblue}
        ER-LAS&$\mathbf{55.3\pm1.6}$ &$\mathbf{25.7\pm0.3}$  &$\mathbf{15.5\pm0.2}$\\
    \bottomrule
    \end{tabular}
\end{table*}

The results in Table~\ref{tab:imbalance mitigation} show that our ER-LAS outperforms all other compared methods for mitigating class imbalance. Next, we will analyze the shortcomings of these methods. ER-CBL re-weights the loss after computing the logits and ground truth, which helps in learning better features of minority classes but fails to eliminate the influence of class-priors to achieve balanced posterior outputs. ER-WN ensures that the model output is not affected by class weight bias. However, we find that CL models are still affected by feature drift\citep{Caccia2021NewIO}, leading to recency bias. Therefore, these two methods cannot truly solve the inter-class imbalance problem in online CL. ER-Up is the closest method to our ER-LAS, but as the inter-class imbalance problem intensifies, it results in a significant computational burden, whereas our method costs almost no additional computational resources. ER-Down, although maintaining inter-class balance during training, discards a majority of valuable incoming training samples. Furthermore, unlike these four methods, our proposed LAS is supported by a statistical ground of underlying class-conditionals.

\subsection{Results on Offline task-IL Scenarios}
\label{subsec:more offline}

In offline task-IL settings, learners have access to a whole dataset for each task and can undergo multiple epochs of training. Previous arts that are highly related to our work have proposed some Logit Rectify methods to alleviate the issue of inter-class imbalance in offline CL and improve learning performance. BiC~\citep{Wu2019LargeSI} adds a bias correction layer to the model and stores a portion of input data as the held-out validation set to calibrate this layer and lessen the model's task-recency bias. E2E~\citep{Castro2018EndtoEndIL} fine-tunes the model with a balanced dataset after each task. IL2M~\citep{Belouadah2019IL2MCI} rescales the model output with historical statistics.  LURIC~\citep{Hou2019LearningAU} combines cosine normalization, less-forget constraint, and inter-class separation to mitigate the adverse effects of class imbalance. We also compare successful offline CL methods (ER~\citep{Chaudhry2019ContinualLW}, DER++~\citep{Buzzega2020DarkEF}, and ER-ACE~\citep{Caccia2021NewIO}). Our experimental settings follow \citep{Buzzega2020DarkEF}. LUCIR fails to work on C-TinyImageNet due to too low memory size. 
{
\begin{table*}[h]
    \caption{Final average accuracy $A_T$ (higher is better) on C-CIFAR10 (5 tasks), C-CIFAR100 (10 tasks), and C-TinyImageNet (10 tasks) in the offline condition. $M=100$. The epoch is set to $50$.}
    \label{tab:offline accuracy}
    \small
    \centering
    \begin{tabular}{@{}l | c | c | c@{}} 
    \toprule
    \textbf{Dataset}         &C-CIFAR10&C-CIFAR100&C-TinyImageNet\\
    \midrule
        BiC             &$23.4$\tiny{$\pm0.8$}  &$15.3$\tiny{$\pm0.1$}  &$10.1$\tiny{$\pm0.1$}\\
        E2E             &$51.6$\tiny{$\pm0.3$}  &$16.7$\tiny{$\pm0.1$}  &$9.0$\tiny{$\pm0.0$}\\ 
        IL2M            &$42.1$\tiny{$\pm0.6$}  &$11.0$\tiny{$\pm0.2$}  &$8.4$\tiny{$\pm0.1$}\\
        LUCIR           &$28.9$\tiny{$\pm1.0$}  &$15.7$\tiny{$\pm0.7$}  &$10.2$\tiny{$\pm0.1$}\\
        ER              &$39.4$\tiny{$\pm0.3$}  &$11.5$\tiny{$\pm0.1$}  &$8.1$\tiny{$\pm0.0$} \\
        DER++           &$55.3$\tiny{$\pm1.2$}  &$14.8$\tiny{$\pm1.8$}  &$9.4$\tiny{$\pm0.3$}\\
        ER-ACE          &$55.9$\tiny{$\pm1.0$}  &$17.7$\tiny{$\pm0.7$}  &$8.7$\tiny{$\pm0.2$}\\
        \rowcolor{myblue}
        ER-LAS          &$53.9$\tiny{$\pm1.0$}  &$16.4$\tiny{$\pm0.2$}  &$10.3$\tiny{$\pm0.1$}\\
        \rowcolor{myblue}
        LwF-LAS         &$\mathbf{57.5}$\tiny{$\mathbf{\pm0.2}$}  &$\mathbf{22.6}$\tiny{$\mathbf{\pm0.1}$} &$\mathbf{12.4}$\tiny{$\mathbf{\pm0.1}$}\\
    \bottomrule
    \end{tabular}
\end{table*}
}

The results in Table~\ref{tab:offline accuracy} demonstrate that ER-LAS still achieves competitive performance in offline CL. When combined with knowledge distillation method LwF~\citep{Li2016LearningWF}, LwF-LAS outperform the other compared methods. These finds indicate that our approach can also effectively mitigate inter-class imbalance and improve performance than previously proposed Logit Rectify methods in offline task-IL setups. Considering the severe impact of recency bias in online CL, our main focus is how to eliminate the adverse effects caused by inter-class imbalance in online settings, and we design a widely applicable LAS algorithm.

\subsection{Comparison to CL Methods Based on Contrastive Learning}
\label{subsec:more contrastive}

\begin{figure}[htb]
    \centering 
    \includegraphics[width=0.8\linewidth]{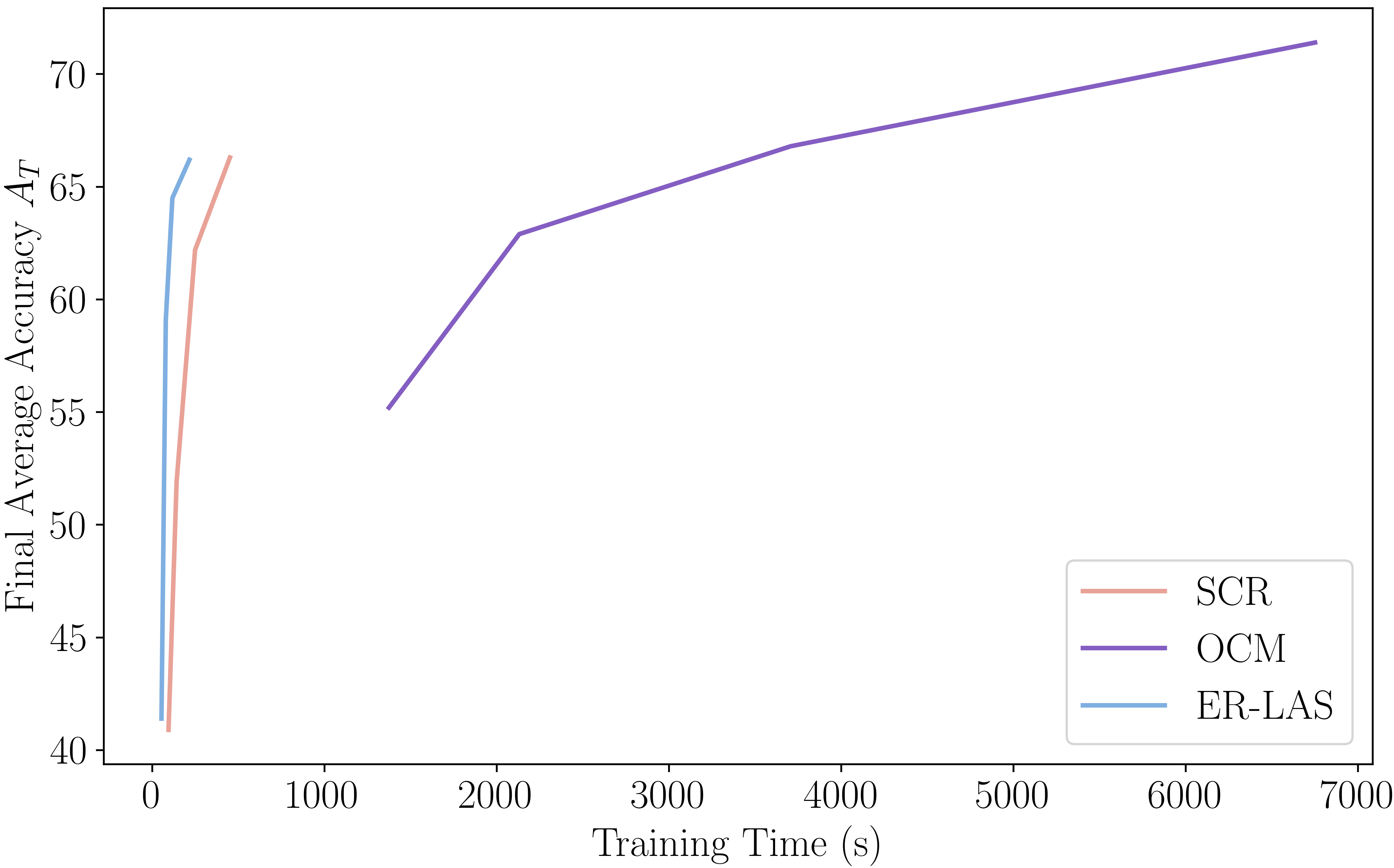}
    \caption{Comparison with online CL methods based on contrastive learning on C-CIFAR10 (5 tasks). Memory size $M=1k$. The $x$-axis represents training time, and the $y$-axis represents the final average accuracy $A_T$ (higher is better). We evaluate the accuracy and the time efficiency of SCR, OCM, and our ER-LAS at batch sizes of 8, 16, 32, and 64. Noting that the time consumption increases as the batch size decreases.
    }
    \label{fig:contrast}
\end{figure}

We compare our method with the online CL methods (SCR~\cite{Mai2021SupervisedCR} and OCM~\cite{Guo2022OnlineCL}) based on contrastive learning. SCR utilizes the NCM classifier and is trained via supervised contrastive learning. OCM employs contrastive learning to maximize mutual information. These methods based on contrastive learning typically require more computational resources, and their performance is influenced by the number of negative samples, but they often achieve better performance. As shown in Figure~\ref{fig:contrast}, we evaluate the training time and the final average accuracy of our ER-LAS and the contrastive learning-based online CL methods under different batch sizes. SCR and OCM require 2x and 30x more computational time than our method, respectively. Although they achieve higher accuracy than our method, LAS exhibits superior overall computational efficiency.

\subsection{Forgetting Rate}
\label{subsec:more forget}

{
\begin{table*}[h]
    \caption{Final average forgetting $F_T$ (lower is better) on C-CIFAR10 (5 tasks), C-CIFAR100 (10 tasks), and C-TinyImageNet (10 tasks). $M$ is the memory buffer size.}
    \label{tab:class cl forget}
    \small
    \centering
    \resizebox{\columnwidth}{!}{
    \begin{tabular}{@{}l | ccc | ccc | ccc@{}} 
    \toprule
    \textbf{Dataset}         &\multicolumn{3}{c|}{C-CIFAR10}&\multicolumn{3}{c|}{C-CIFAR100}& \multicolumn{3}{c}{C-TinyImageNet}\\
    \midrule
    \textbf{Method}          & $M=0.5k$ & $M=1k$ & $M=2k$   & $M=0.5k$ & $M=1k$ & $M=2k$   & $M=0.5k$ & $M=1k$ & $M=2k$ \\
    % \midrule
    \midrule
        ER              &$43.0$\tiny{$\pm1.5$}&$36.2$\tiny{$\pm1.6$}&$24.5$\tiny{$\pm1.3$}  &$31.1$\tiny{$\pm0.6$}&$23.2$\tiny{$\pm1.0$}&$23.2$\tiny{$\pm0.6$}  &$38.5$\tiny{$\pm0.5$}&$33.4$\tiny{$\pm0.3$}&$27.8$\tiny{$\pm0.3$}\\
        DER++           &$29.3$\tiny{$\pm1.2$}&$31.6$\tiny{$\pm2.9$}&$32.4$\tiny{$\pm2.3$}  &$37.6$\tiny{$\pm0.5$}&$34.5$\tiny{$\pm0.5$}&$36.4$\tiny{$\pm0.6$}   &$38.6$\tiny{$\pm0.2$}&$37.2$\tiny{$\pm0.3$}&$37.2$\tiny{$\pm0.2$}\\
        MRO             &$26.1$\tiny{$\pm1.5$}&$21.0$\tiny{$\pm1.2$}&$8.9$\tiny{$\pm0.8$} &$13.5$\tiny{$\pm0.3$}&$9.3$\tiny{$\pm0.3$}&$6.3$\tiny{$\pm0.2$} &$\mathbf{11.1}$\tiny{$\mathbf{\pm0.1}$}&$10.9$\tiny{$\pm0.2$}&$8.4$\tiny{$\pm0.2$}\\
        SS-IL           &$22.0$\tiny{$\pm0.8$}&$20.0$\tiny{$\pm0.9$}&$16.5$\tiny{$\pm0.5$}  &$11.8$\tiny{$\pm0.3$}&$10.0$\tiny{$\pm0.2$}&$8.1$\tiny{$\pm0.3$}   &$14.5$\tiny{$\pm0.2$}&$12.2$\tiny{$\pm0.2$}&$10.0$\tiny{$\pm0.8$}\\
        CLIB            &$30.4$\tiny{$\pm1.6$}&$17.7$\tiny{$\pm1.5$}&$16.1$\tiny{$\pm1.3$}  &$25.9$\tiny{$\pm0.3$}&$14.9$\tiny{$\pm0.3$}&$7.6$\tiny{$\pm0.3$}   &$30.1$\tiny{$\pm0.3$}&$20.4$\tiny{$\pm0.3$}&$10.8$\tiny{$\pm0.3$}\\
        ER-ACE          &$\mathbf{11.0}$\tiny{$\mathbf{\pm1.6}$}&$\mathbf{16.1}$\tiny{$\mathbf{\pm1.3}$}&$10.1$\tiny{$\pm1.3$}  &$\mathbf{9.3}$\tiny{$\mathbf{\pm0.7}$}&$\mathbf{7.9}$\tiny{$\mathbf{\pm0.5}$}&$\mathbf{5.6}$\tiny{$\mathbf{\pm0.7}$}  &$13.8$\tiny{$\pm0.3$}&$\mathbf{9.9}$\tiny{$\mathbf{\pm0.3}$}&$\mathbf{7.5}$\tiny{$\mathbf{\pm0.4}$}\\
        ER-OBC    &$37.3$\tiny{$\pm0.8$}&$19.5$\tiny{$\pm0.6$}&$30.5$\tiny{$\pm0.8$} &$24.0$\tiny{$\pm0.3$}&$22.4$\tiny{$\pm0.4$}&$18.7$\tiny{$\pm0.4$} &$36.2$\tiny{$\pm0.2$}&$29.4$\tiny{$\pm0.2$}&$21.9$\tiny{$\pm0.2$}\\
        \rowcolor{myblue}
        ER-LAS          &$28.5$\tiny{$\pm1.3$}&$18.5$\tiny{$\pm1.4$}&$\mathbf{7.1}$\tiny{$\mathbf{\pm1.1}$}   &$22.1$\tiny{$\pm0.4$}&$11.5$\tiny{$\pm0.6$}&$9.3$\tiny{$\pm0.7$}  &$26.9$\tiny{$\pm0.2$}&$19.9$\tiny{$\pm0.4$}&$10.3$\tiny{$\pm0.2$}\\
    \bottomrule
    \end{tabular}}
\end{table*}
}

We compare the forgetting rate of each method on C-CIFAR10, C-CIFAR100, and C-TinyImageNet in Table~\ref{tab:class cl forget}. In most settings, ER-ACE achieved the lowest forgetting rate, except for when compared to our proposed ER-LAS on C-CIFAR10 with $M=2k$, and to MRO on C-TinyImageNet with $M=0.5k$. Noting that the lowest forgetting rate does not necessarily correspond to the highest accuracy. Moreover, remarkable reductions in the forgetting rate can be achieved by deliberately adjusting the hyperparameters of our method, but at the cost of accuracy. Currently, our method achieves the optimal stability-plasticity trade-off.

\subsection{Prediction Results on C-ImageNet}
\label{subsec:more predict}

\begin{figure}[htb]
    \centering 
    \includegraphics[width=0.99\columnwidth]{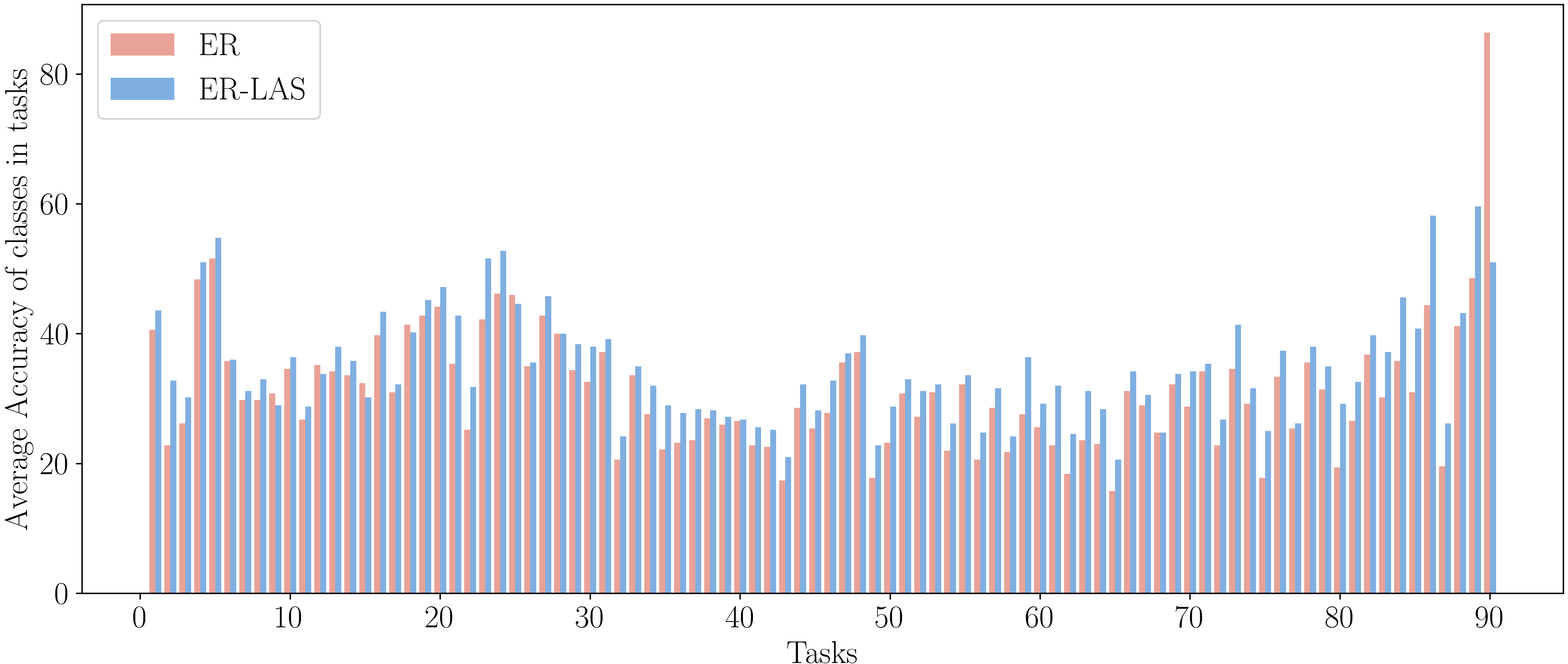}
    \caption{Prediction results by ER and ER-LAS on C-ImageNet (90 tasks). We calculate the average accuracy of classes within each task to demonstrate the recency bias.
    }
    \label{fig:pred imagenet}
\end{figure}

We present the prediction results of ER and our proposed ER-LAS on C-ImageNet after training, as shown in Figure~\ref{fig:pred imagenet}. Recalling that ER assigns 38\% of the test samples to the most recently learned classes in C-CIFAR100 (Figure~\ref{fig:er-las} in \S\ref{sec:method}). ER also outperforms ER-LAS on the last task, but is inferior to ER-LAS on all other tasks. This is due to the larger task sequence and the more total number of classes in C-ImageNet than in C-CIFAR100, resulting in a much more severe recency bias for the ER method. However, our ER-LAS successfully eliminates the recency bias as expected, and as a result, achieves a remarkably lower forgetting rate and the highest accuracy in the experiments of \S\ref{subsec:online class-il cl results}. These results validate that inter-class imbalance is more severe in long sequential tasks and demonstrate that our method can adapt to learning from such highly imbalanced data streams by pursuing the class-conditional function.

\subsection{Class-balanced Accuracy on C-iNaturalist}

{
\begin{table}[h]
\centering
\caption{Final average accuracy $A_T$ and final average class-balanced accuracy $A_T^{\text{cbl}}$ (both higher is better) on C-iNaturalist (26 tasks). We show the results of top-3 methods. Memory sizes are $M=20k$.}
\label{tab:inat cbl}
\begin{tabular}{@{}l|cc@{}}
\toprule
\textbf{Dataset}  & \multicolumn{2}{c}{iNaturalist} \\
\midrule
\textbf{Method}   & $A_T \uparrow$      & $A_T^{\text{cbl}} \uparrow$ \\
\midrule
ER                & $4.66$\tiny{$\pm0.01$} & $6.25$\tiny{$\pm0.01$}  \\
ER-ACE            & $5.68$\tiny{$\pm0.01$} & $6.32$\tiny{$\pm0.01$}  \\ 
MRO               & $4.96$\tiny{$\pm0.0$}  & $4.47$\tiny{$\pm0.01$}  \\
\rowcolor{myblue}
ER-LAS            & $\bf8.11$\tiny{$\bf\pm0.01$} & $\bf8.62$\tiny{$\bf\pm0.01$}  \\
\bottomrule
\end{tabular}%
\end{table}
}

As we aim to pursue the {optimal} classifier that minimizes the class-balanced error on imbalanced data streams, we also evaluate the class-balanced accuracy of our method and baselines on C-iNaturalist. As shown in Table~\ref{tab:inat cbl}, ER-LAS$^\dagger$ achieves the best performance in both accuracy and class-balanced accuracy, validating the effectiveness of our optimization towards {this} optimal estimator.

\end{document}